\documentclass[submission,copyright,creativecommons]{eptcs}
\providecommand{\event}{ICLP 2024}

\usepackage[utf8]{inputenc}
\usepackage{amsmath}
\usepackage{amssymb}
\usepackage{microtype}
\usepackage{listings}
\usepackage{graphicx}
\usepackage{xcolor}

\usepackage{url}

\providecommand{\Underscore}{\textunderscore}

\lstdefinelanguage{clingo}{basicstyle=\ttfamily,keywordstyle=[1]\bfseries,keywordstyle=[2]\bfseries,keywordstyle=[3]\bfseries,showstringspaces=false,literate={_}{\Underscore}1 {\%\%}{}0,escapeinside={\#(}{\#)},alsoletter={\#,\&},keywords=[1]{not,from,import,def,if,else,elif,return,while,break,and,or,for,in,del,and,class,with,as,is,yield,async},keywords=[2]{\#const,\#show,\#minimize,\#base,\#theory,\#count,\#external,\#program,\#script,\#end,\#heuristic,\#edge,\#project,\#show,\#sum},keywords=[3]{&,&dom,&sum,&diff,&show},morecomment=[l]{\#\ },morecomment=[l]{\%\ },morestring=[b]",stringstyle={\itshape},commentstyle={\color{darkgray}}}

\lstdefinelanguage{python}{basicstyle=\ttfamily,keywordstyle=[1]\bfseries,showstringspaces=false,literate={_}{\Underscore}{1},escapeinside={\#(}{\#)},alsoletter={\#,\&},keywords=[1]{not,from,import,def,if,else,elif,return,while,break,and,or,for,in,del,and,class,with,as,is,yield,async},morecomment=[l]{\#\ },morestring=[b]",stringstyle={\itshape},commentstyle={\color{darkgray}}}

 \providecommand{\sysfont}{\textit}

\newcommand{\clorm}{\sysfont{clorm}}
\newcommand{\clingraph}{\sysfont{clingraph}}
\newcommand{\clinguin}{\sysfont{clinguin}}
\newcommand{\Clinguin}{\sysfont{Clinguin}}

\newcommand{\clasp}{\sysfont{clasp}}

\newcommand{\clingo}{\sysfont{clingo}}
\newcommand{\clingodl}{\clingoM{dl}}

\newcommand{\wasp}{\sysfont{wasp}}

\newcommand{\clingoM}[1]{\clingo{\small\textnormal{[}\textsc{#1}\textnormal{]}}}

\newcommand{\prolog}{Prolog}
\newcommand{\python}{Python}

 \newcommand{\domainstate}[0]{\textit{domain-state}}
\newcommand{\uistate}[0]{\textit{ui-state}}
\newcommand{\domctl}[0]{\textit{domain-control}}
\newcommand{\uictl}[0]{\textit{ui-control}}

\lstdefinelanguage{clingos}{language=clingo,basicstyle=\small\ttfamily }

\makeatletter\lst@AddToHook{OnEmptyLine}{\vspace{\dimexpr-\baselineskip+\medskipamount\relax}}\makeatother

\usepackage{iftex}

\ifpdf
  \usepackage{underscore}         \usepackage[T1]{fontenc}        \else
  \usepackage{breakurl}           \fi

\title{ASP-driven User-interaction with \Clinguin}
\author{Alexander Beiser
\and
Susana Hahn
\and
Torsten Schaub
\and
\institute{University of Potsdam, Germany \quad Potassco Solutions, Germany}
\email{alexander.beiser@tuwien.ac.at  \quad hahnmartinlu@uni-potsdam.de \quad torsten@uni-potsdam.de}
}

\newcommand{\titlerunning}{ASP-driven User-interaction with \Clinguin}
\newcommand{\authorrunning}{A. Beiser, S. Hahn \& T. Schaub}

\hypersetup{
  bookmarksnumbered,
  pdftitle    = {\titlerunning},
  pdfauthor   = {\authorrunning},
  pdfsubject  = {EPTCS},               }

\begin{document}

\maketitle

\begin{abstract}
  We present \clinguin, a system for ASP-driven user interface design.
  \Clinguin\ streamlines the development of user interfaces for ASP developers
  by letting them build interactive prototypes directly in ASP,
  eliminating the need for separate frontend languages.
  To this end,
  \clinguin\ uses a few dedicated predicates to define user interfaces and the treatment of user-triggered events.
  This simple design greatly facilitates the specification of user interactions with an ASP system,
  in our case \clingo.
\end{abstract}
 \section{Introduction}\label{sec:introduction}

The growing popularity of Answer Set Programming (ASP;~\cite{lifschitz19a})
in both academia and industry necessitates the development of user-friendly graphical interfaces to cater to end users.
This is especially critical for interactive applications
where users engage in iterative feedback loops with ASP systems.
Examples include timetabling or product configuration tools.
This leads to challenges in frontend development
and requires skills in areas beyond ASP development.
In addition, custom solutions have a limited reach, as they cannot be easily adapted.

\Clinguin\ addresses this challenge and streamlines User Interface (UI) development for ASP developers
by letting them build interactive prototypes directly in ASP,
eliminating the need for separate frontend languages.
To this end,
\clinguin\ uses a few dedicated predicates to define UIs and the treatment of user-triggered events.
This simple design greatly facilitates the specification of user interactions with an ASP system,
in our case \clingo~\cite{karoscwa21a}.
Our approach shares similarities with the ASP-driven visualization system \clingraph~\cite{hasascst22a}.
In fact, \clinguin\ can be regarded as the interactive
counterpart of \clingraph,
whose single-shot approach lacks any interaction capabilities.

In what follows,
we rely on a basic acquaintance with ASP, the ASP system \clingo~\cite{PotasscoUserGuide},
and some rudimentary \python\ knowledge.
We explain specialized concepts as they are introduced throughout the text.

\section{Architecture and Workflow of \clinguin}
\label{sec:workflow}

\Clinguin\ uses a client-server architecture, where communication occurs via the HTTP protocol
(RESTful\footnote{\url{https://en.wikipedia.org/wiki/REST}}).
JSON is used for message content between client and server.
The server leverages \clingo~(5.7) and calculates the information needed to build the UI;
this is used by the client to render the corresponding UI and
update it based on the user's interaction.
The update is either handled directly at the client-side or sent back to the server for further processing,
triggering a new interaction loop.

\begin{figure}[!h]
    \includegraphics[width=\textwidth]{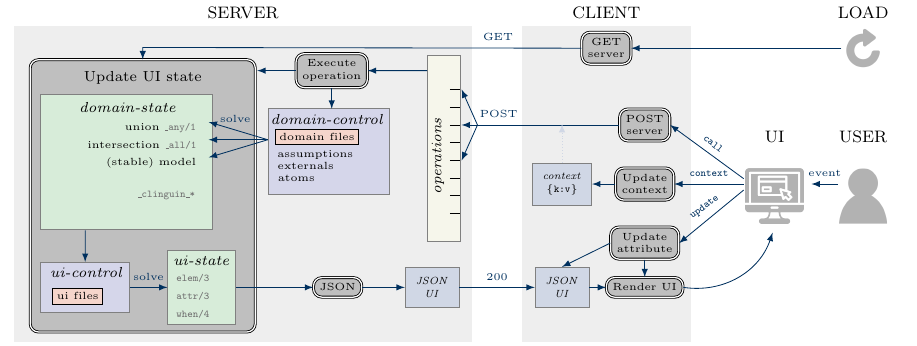}
    \caption{Architecture and Workflow of \clinguin}
    \label{fig:clinguin}
\end{figure}
Figure~\ref{fig:clinguin} illustrates the architecture and workflow of \clinguin.
It focuses on the key technical components: server, client, and user interface (UI).
Furthermore, it shows how the UI is generated and how user interactions are handled within the system.

In what follows,
we illustrate \clinguin's approach with a detailed, step-by-step walkthrough using a simple running example
about assigning people to seats where a cat person and a dog person cannot share the same table.
The corresponding problem instance and encoding are given in Listings~\ref{prg:instance} and~\ref{prg:encoding}, respectively.
We refer to them as the \emph{domain files} among the input files; they are highlighted in pink in Figure~\ref{fig:clinguin}.
\begin{lstlisting}[caption={A simple problem instance (\texttt{ins.lp})},label={prg:instance},language=clingos]
person("Susana",cat). person("Alexander",cat). person("Torsten",dog).
seat((1,1..2)). seat((2,1..3)).
\end{lstlisting}
\begin{lstlisting}[caption={A simple problem encoding (\texttt{enc.lp})},label={prg:encoding},language=clingos]
{assign(P,S):seat(S)}:- person(P,_).%%#(\label{enc:ch}#)

:- person(P,_), #count{S:assign(P, S)}!=1, cons(exacly_one, _). %%#(\label{enc:one}#)
:- assign(P,S), assign(P',S), P'>P, cons(only_one, _).
:- assign(P,(T,_)), assign(P',(T,_)), person(P,cat), person(P',dog),
   cons(different_type, _).%%#(\label{enc:tri}#)

cons(exacly_one, "All people need exactly one seat").%%#(\label{c:one}#)
cons(only_one, "Two people can not be seated on the same seat").
cons(different_type, "All people on a table must prefer the same pet").%%#(\label{c:tri}#)
\end{lstlisting}
The choice in Line~\ref{enc:ch} of Listing~\ref{prg:encoding} generates possible assignments of a person $p$ to a seat $s$.
Each seat $s$ is defined by a pair $(t,c)$ of a numbered table $t$ and chair $c$.
Lines~\ref{enc:one} to~\ref{enc:tri} constrain the assignment, ensuring that a person is assigned to exactly one seat
where everyone at the table shares the same pet preference.
Each of these integrity constraints is conditioned with an instance of \lstinline{cons/2} defined as facts in Lines~\ref{c:one} to~\ref{c:tri}.
The first argument of \lstinline{cons/2} identifies the constraint violation,
while the second argument provides the corresponding user-friendly explanation.
Although the introduction of these atoms may appear unnecessary at this point,
they will be used to identify constraint violations in Section~\ref{sec:extensibility}.

The server is started with the domain and UI files as command-line arguments:
\begin{lstlisting}[basicstyle=\small\ttfamily,numbers=none]
clinguin server --domain-files ins.lp enc.lp --ui-files ui-tables.lp
\end{lstlisting}
Using the domain files, the server creates the \clingo\ object \domctl\ (in purple in Figure~\ref{fig:clinguin}),
which employs multi-shot solving by default.
The other input files, called \emph{ui-files} and given in pink in Figure~\ref{fig:clinguin},
are used to generate
a single stable model composed of
atoms defining the layout, style, and functionality of the interface,
collectively forming the \uistate.
To this end, another \clingo\ object, \uictl\ in purple in Figure~\ref{fig:clinguin}, is restarted on each update.

Upon launching the client with `\lstinline{clinguin client}',
it requests the UI state (or \uistate) from the server (see the \textit{GET} arrow in Figure~\ref{fig:clinguin}).
Once received, the client utilizes a front-end language to render the corresponding user interface.
We use
\emph{Angular}\footnote{\url{https://angular.io}} as front-end language with
\emph{Bootstrap}\footnote{\url{https://getbootstrap.com}} as a toolkit for customization.
Subsequent user interactions with the UI (usually) generate new requests to the server,
providing details about the selected operations.
These operations are predefined by the server and
allow users to interact with the \domctl\ in different ways.
Examples include
adding a user selection as an assumption to the solver,
setting the value of an external atom, or
obtaining the next solution\footnote{
  For the full list of supported elements and operations, we refer the reader to the system's documentation at \url{https://clinguin.readthedocs.io/en/latest}.
  \label{doc}
}.
Once the server completes the selected operations,
it constructs a hierarchical JSON structure of the updated \uistate\ and
returns it to the client for rendering.

The \uistate\ is defined by predicates \lstinline|elem/3|, \lstinline|attr/3| and \lstinline|when/4|,
for specifying the UI's layout, style and functionality, respectively
(see green box with header \uistate\ in Figure~\ref{fig:clinguin}).
The corresponding atoms are mapped into Python classes using \clorm\footnote{\url{https://github.com/potassco/clorm}},
a Python library that provides an object-relational mapping interface to \clingo.
An atom \lstinline|elem(X,T,X')| defines an element \lstinline|X| of type \lstinline|T| inside element \lstinline|X'|.
Such UI \emph{elements} are the visual representations of objects or features in an interface,
such as a button, text field, dropdown menu, and more\textsuperscript{\ref{doc}}.
The \emph{attributes} of an element, such as position and style, are specified by \lstinline|attr(X,K,V)|,
where \lstinline|X| is an element, and \lstinline|K| and \lstinline|V| denote the attribute's name (key) and value, respectively.

The reactive behavior of the UI is defined by atoms of form \lstinline|when(X,E,A,P)|,
which can be interpreted as expressing:
``When event \lstinline|E| is triggered on element \lstinline|X|,
it is followed by an action \lstinline|A| with arguments \lstinline|P|''.
An \emph{event} refers to an action initiated by the user, such as clicking, double-clicking, or entering text.
An \emph{action} is a system response triggered by the UI event.
This action must be one of the following:
\begin{itemize}
\item a \lstinline|call| to the server (POST operation),
  where \lstinline|P| represents one or multiple server operations,
\item a local attribute \lstinline|update|,
  where \lstinline|P| is a triple \lstinline|(X',K,V)|
  leading to an update of attribute \lstinline|K| on element \lstinline|X'| by \lstinline|V|,
\item a local update to a \lstinline|context| defined as a dictionary,
  where \lstinline|P| is the key-value pair to update and keep as local memory.
\end{itemize}

The atoms constituting the \uistate\ are generated from the encodings provided as ui-files
along with facts describing the \domainstate\
(see green box with the identical header in Figure~\ref{fig:clinguin}).
The latter state provides valuable insights into the current state of the \domctl\ object,
including
the intersection and union of the stable models generated by the \domctl,
a single stable model for focused exploration, and
internal information encapsulated in atoms whose predicates typically start with \lstinline|_clinguin_|.
Collectively, these facts represent relevant information for generating the UI.

\begin{figure}[ht]
    \centering
    \frame{\includegraphics[scale=0.20]{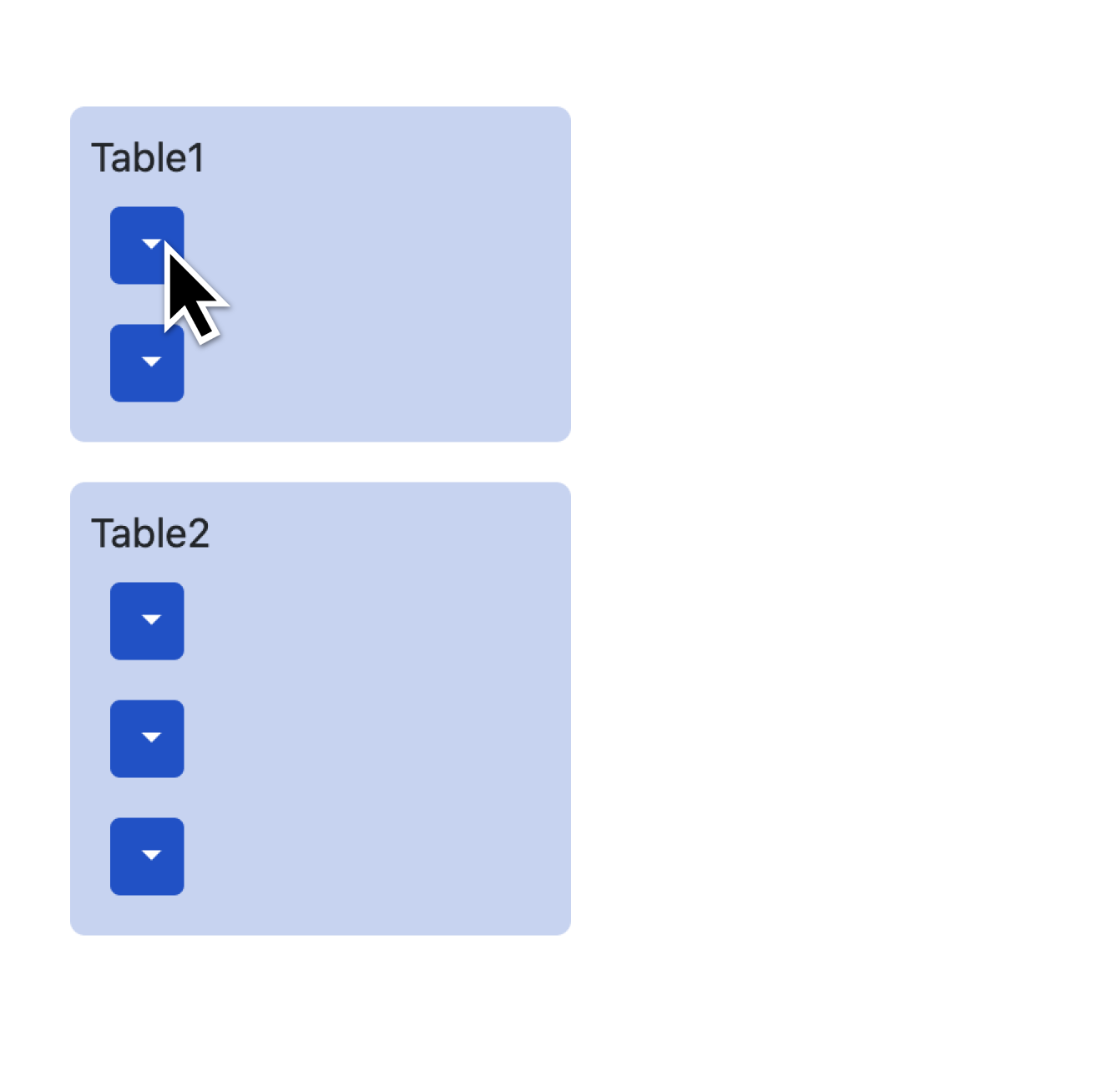}}
    \frame{\includegraphics[scale=0.20]{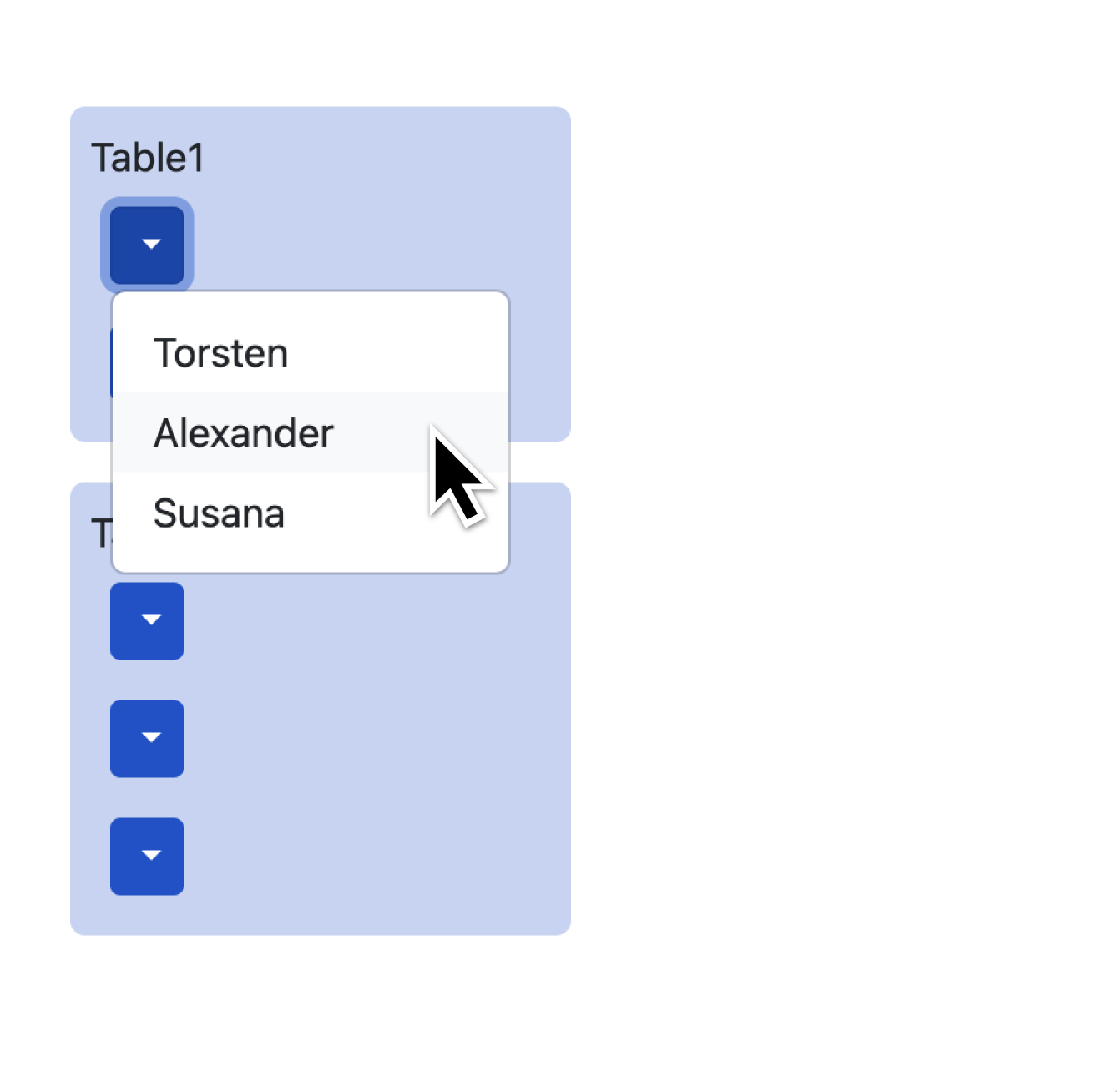}}
    \frame{\includegraphics[scale=0.20]{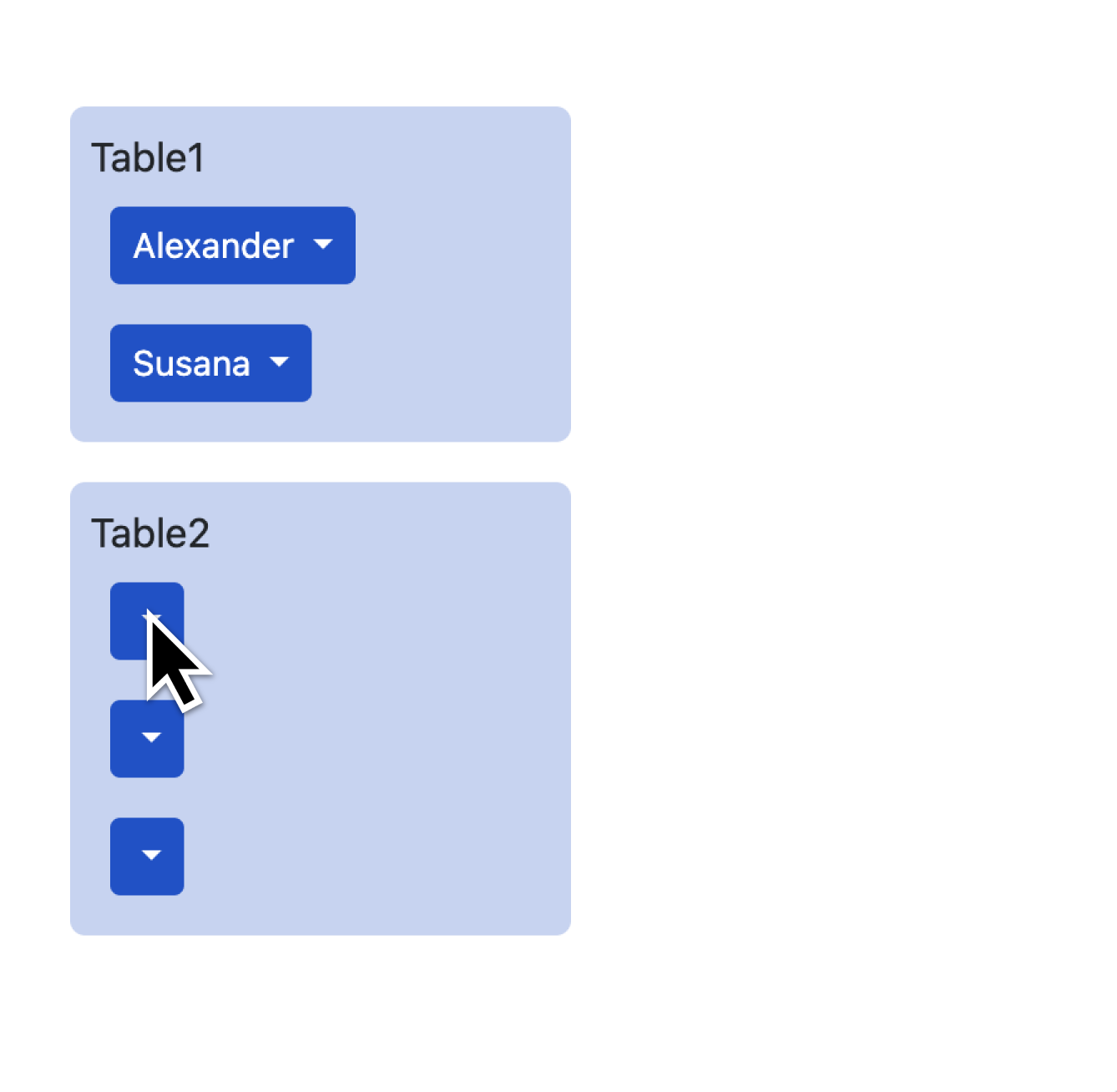}}
    \frame{\includegraphics[scale=0.20]{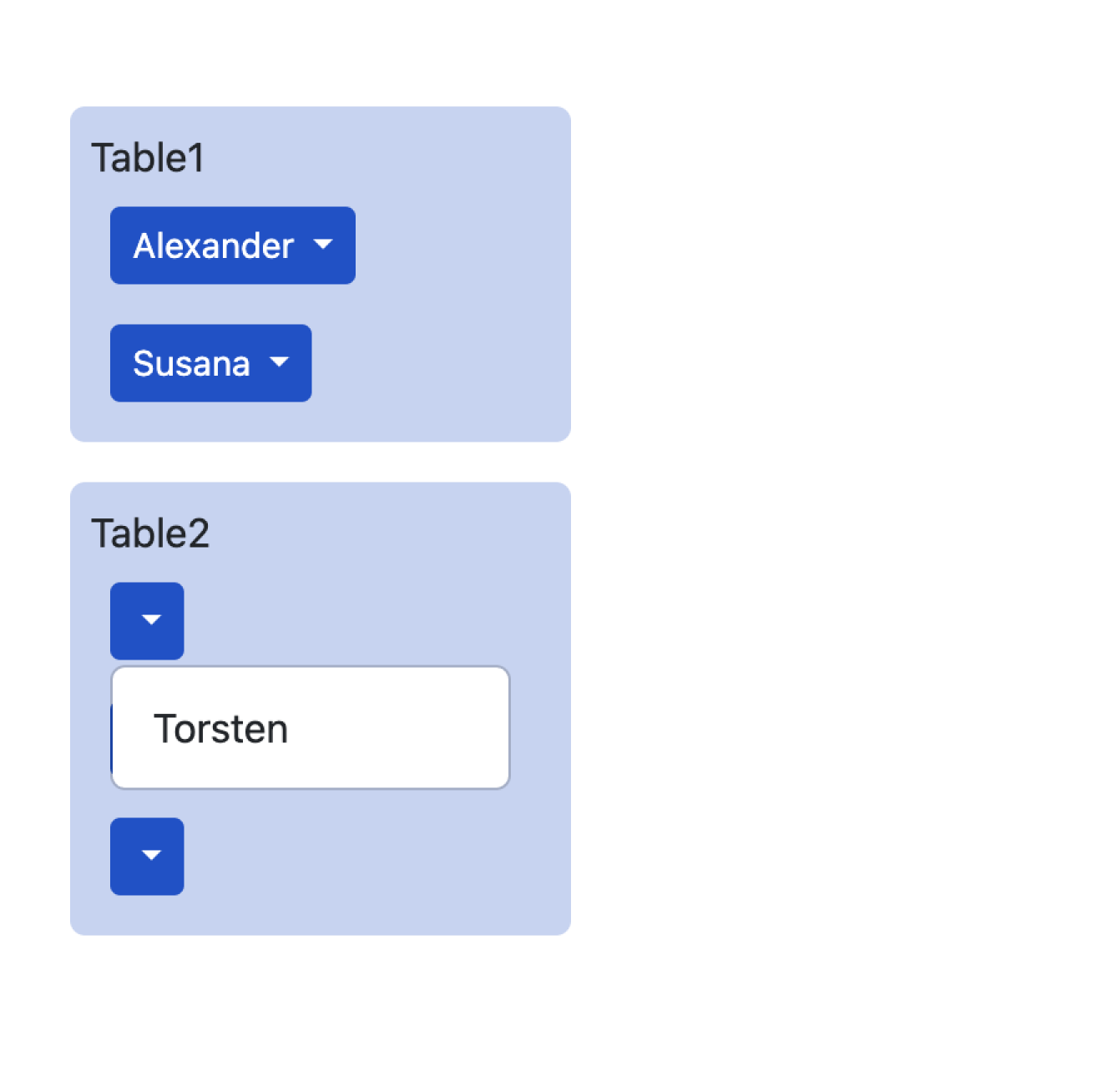}}
    \caption{User interaction via mouse actions using ui-file \texttt{ui-tables.lp}.}
    \label{fig:table}
\end{figure}

Let us illustrate this with our example.
Listing~\ref{prg:uitables} shows the ui-file generating the UI snapshots in Figure~\ref{fig:table}.
\begin{lstlisting}[float=ht,caption={An encoding for table handling (\texttt{ui-tables.lp})},label={prg:uitables},language=clingos]
elem(w, window, root).%%#(\label{ui:tables:elem:one}#)
attr(w, flex_direction, row).%%#(\label{ui:tables:attr:one}#)

elem(tables, container, w).%%#(\label{ui:tables:container:one}#)

elem(table(T), container, tables):- seat((T,_)).%%#(\label{ui:tables:container:two}#)
attr(table(T), order, T):- seat((T,_)).%%#(\label{ui:tables:attr:two}#)
attr(table(T), width, 200):- seat((T,_)).%%#(\label{ui:tables:attr:tri}#)
attr(table(T), class, ("bg-primary";"bg-opacity-25";"rounded";%%#(\label{ui:tables:class:one:a}#)
                       "d-flex";"flex-column";"align-items-start";%%#(\label{ui:tables:class:one:b}#)
                       "p-2";"m-2"%%#(\label{ui:tables:class:one:c}#)
                     )):- seat((T,_)).%%#(\label{ui:tables:class:one:d}#)

elem(table_label(T), label, table(T)):- seat((T,_)).%%#(\label{ui:tables:label:one}#)
attr(table_label(T), order, 1):- seat((T,_)).
attr(table_label(T), label, @concat("Table",T)):- seat((T,_)).%%#(\label{ui:tables:label:tri}#)

elem(seat_dd((T,C)), dropdown_menu, table(T)):- seat((T,C)).%%#(\label{ui:tables:dropdown:one:a}#)
attr(seat_dd((T,C)), order, C+1):- seat((T,C)).%%#(\label{ui:tables:dropdown:one:b}#)
attr(seat_dd(S), class, ("btn-sm";"btn-primary";"m-2")):- seat(S).%%#(\label{ui:tables:dropdown:one:c}#)
attr(seat_dd(S), selected, P):- _all(assign(P,S)).%%#(\label{ui:tables:dropdown:one:d}#)

elem(seat_ddi(S,P), dropdown_menu_item, seat_dd(S)):-_any(assign(P,S)).%%#(\label{ui:tables:dropdown:two:a}#)
attr(seat_ddi(S,P), label, P):- _any(assign(P,S)).%%#(\label{ui:tables:dropdown:two:b}#)
when(seat_ddi(S,P), click, call, add_assumption(assign(P,S), true)):- %%#(\label{ui:tables:dropdown:two:c}#)
                     _any(assign(P,S)).
\end{lstlisting}
Line~\ref{ui:tables:elem:one}
creates a window element labeled \lstinline|w| and
places it inside the overarching root element.
Line~\ref{ui:tables:attr:one} adds an attribute to window \lstinline|w|
stating that its children elements form a row.
Similarly,
Line~\ref{ui:tables:container:one}
creates a container \lstinline|tables| and
places it inside window \lstinline|w|.
This container groups all elements representing tables.
Accordingly,
Line~\ref{ui:tables:container:two} defines a container for each table \lstinline|table(T)| with number \lstinline|T|.
The table numbers are drawn from the instance using atoms with predicate \lstinline|seat/1|
(as seen in Listing~\ref{prg:instance}).
Line~\ref{ui:tables:attr:two} uses attribute \lstinline|order| to order tables based on their number.
Similarly, Line~\ref{ui:tables:attr:tri} sets the width of these elements.

In HTML, the \emph{class} attribute specifies one or more class names for an element.
These class names correspond to styles defined in a style sheet,
which determines the visual properties of the element.
\emph{Bootstrap} provides a predefined set of classes that help ensure consistent styling
across a user interface.
As common in UI design,
these classes draw from a custom color palette.
For \clinguin, we crafted this color palette with primary (blue) and secondary (purple) colors
as well as special colors representing information, warnings and errors.
Lines~\ref{ui:tables:class:one:a} to~\ref{ui:tables:class:one:d}
set eight of such \emph{Bootstrap} \lstinline|class| names for each table element.
This is done via \clingo's pooling operation `\lstinline|;|' to expand rules.
In detail,
the \lstinline|class| names in Line~\ref{ui:tables:class:one:a}
address the background color, opacity and rounded corners.
Line~\ref{ui:tables:class:one:b} deals with the orientation of all elements in the container.
And Line~\ref{ui:tables:class:one:c} addresses padding and margin.

Lines~\ref{ui:tables:label:one} to~\ref{ui:tables:label:tri}
create a label element with the table number as title in the first position of the table container.
To facilitate this,
\clinguin\ provides several external \python\ functions,
such as \lstinline|@concat| for label assembly Line~\ref{ui:tables:label:tri}.

So far, we only used ASP to generate a set of facts capturing static aspects of a user interface.
Next,
we want to leverage ASP to present users with a well-defined set of choices for selection.
To this end,
we differentiate between necessary and possible selections in view of what is already chosen by the user.
These selections can be captured via the intersection and union of the stable models
of the domain files,
while also incorporating the user's selections.
A necessary selection belongs to all stable models, and
a possible one to at least one.
In technical terms,
this is achieved by manipulating and reasoning with the ASP system \clingo\ encapsulated within the \domctl\ object.
Recall that the latter is initialized with the domain files.
User selections can alternatively be incorporated into the \domctl\ object in terms of
assumptions, externals, or regular atoms~(cf.~\cite{karoscwa21a}).
To further include atoms belonging to the intersection and union of the stable models, we reify them via predicates \lstinline|_all/1| and \lstinline|_any/1|, respectively, and
add the resulting atoms to the current \domainstate.
This is accomplished by two consecutive invocations to the current \domctl\ object,
once setting \clingo\ option \texttt{--enum-mode} to \texttt{cautious} and then to \texttt{brave}.
Although predicates \lstinline|_all/1| and \lstinline|_any/1| resemble epistemic operators,
their usage is restricted to passing information from the domain to the UI side.

Now, let us explore how \clinguin\ manages this in our running example.
Lines~\ref{ui:tables:dropdown:one:a} to~\ref{ui:tables:dropdown:one:d} define
a \lstinline|dropdown_menu| for each seat and
add it to the corresponding \lstinline|table| container.
The heading text of each menu is defined with attribute \lstinline|selected|.
We use this attribute to indicate necessary selections in Line~\ref{ui:tables:dropdown:one:d}.
Only if a person~\lstinline|P| is assigned the same seat~\lstinline|S| in all stable models,
expressed by \lstinline|_all(assign(P,S))|,
its name is shown as the respective menu text.
Similarly,
Lines~\ref{ui:tables:dropdown:two:a} to~\ref{ui:tables:dropdown:two:c}
define all possible seat selections by
\lstinline|dropdown_menu_item|[s] in terms of \lstinline|_any(assign(P,S))|.

Line~\ref{ui:tables:dropdown:two:c} dictates the actions occurring
when a user \lstinline|click|[s] on an item within the dropdown menu.
If so,
a \lstinline|call| action is initiated and transmitted to the server.
In our case,
we model user selections as assumptions.
Accordingly, server operation \lstinline|add_assumption| is invoked with arguments \lstinline|assign(p,s)| and \lstinline|true|,
reflecting the user's selection of person~\lstinline|p| at seat~\lstinline|s|.
This operation results in the addition of atom \lstinline|assign(p,s)| to the \domctl\ object as a true assumption.\footnote{
  As discussed below, assumptions are recorded via predicate \texttt{_clinguin_assume/2} in the \domainstate.
}

Semantically,
this amounts to adding the integrity constraint `\lstinline|:- not assign(p,s).|' and
thus forcing the domain encoding to infer \lstinline|assign(p,s)|.
All this is reflected by the upper path from the UI to \domctl\ in Figure~\ref{fig:clinguin}.

The second screenshot in Figure~\ref{fig:table} shows
that initially all persons are possible selections for the first seat at Table~1.
Once Alexander's seat is chosen, \clinguin\ adds the corresponding assumption to the \domctl\ object.
This automatically leads to Susana's assignment to the second seat,
since all resulting stable models agree on this.
Moreover, all three seats at Table~2 have now only the single option Torsten left,
since none of the remaining stable models seats Alexander or Susana at this table.

\begin{figure}[ht]
    \centering
    \frame{\includegraphics[scale=0.20]{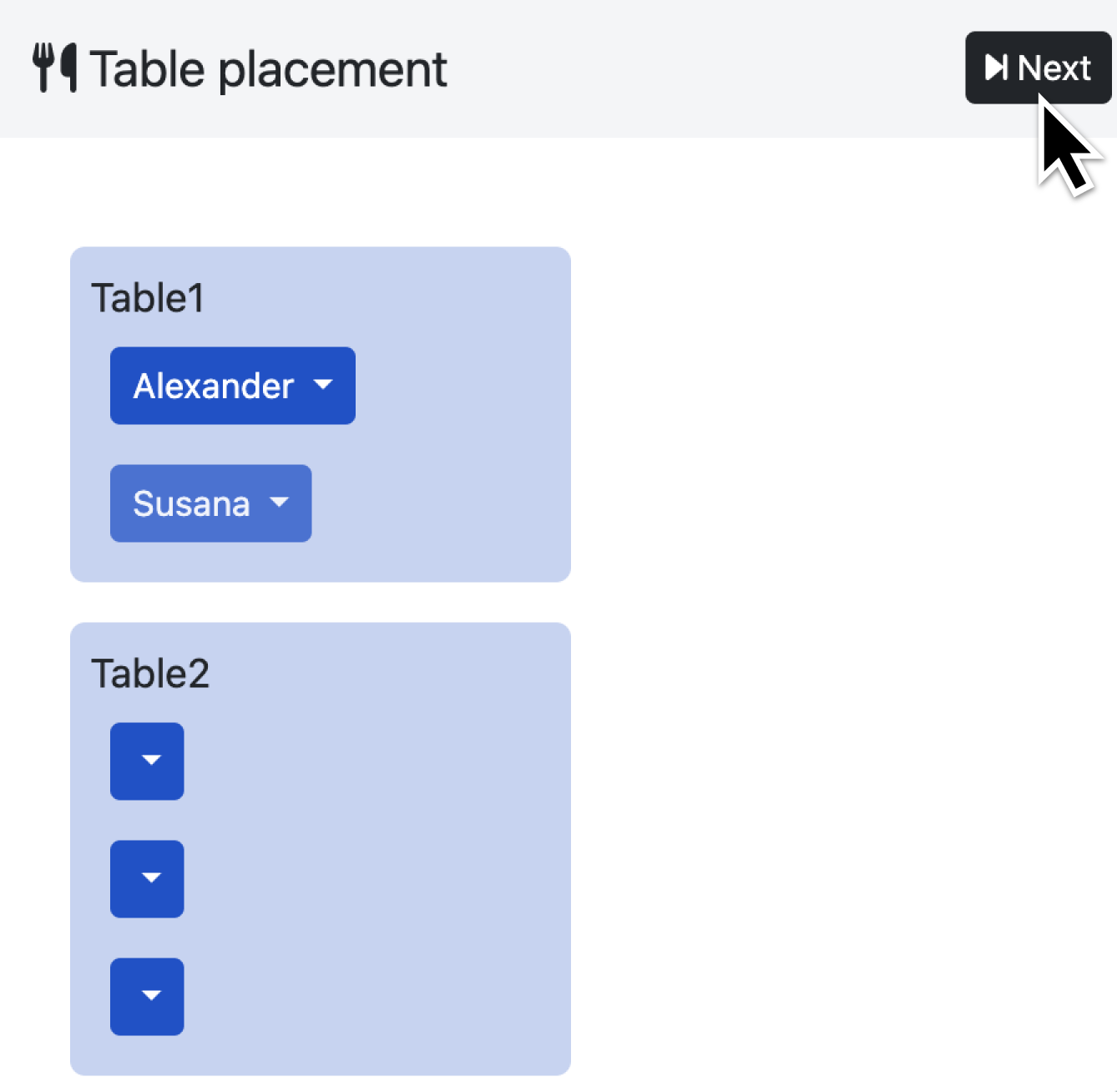}}
    \frame{\includegraphics[scale=0.20]{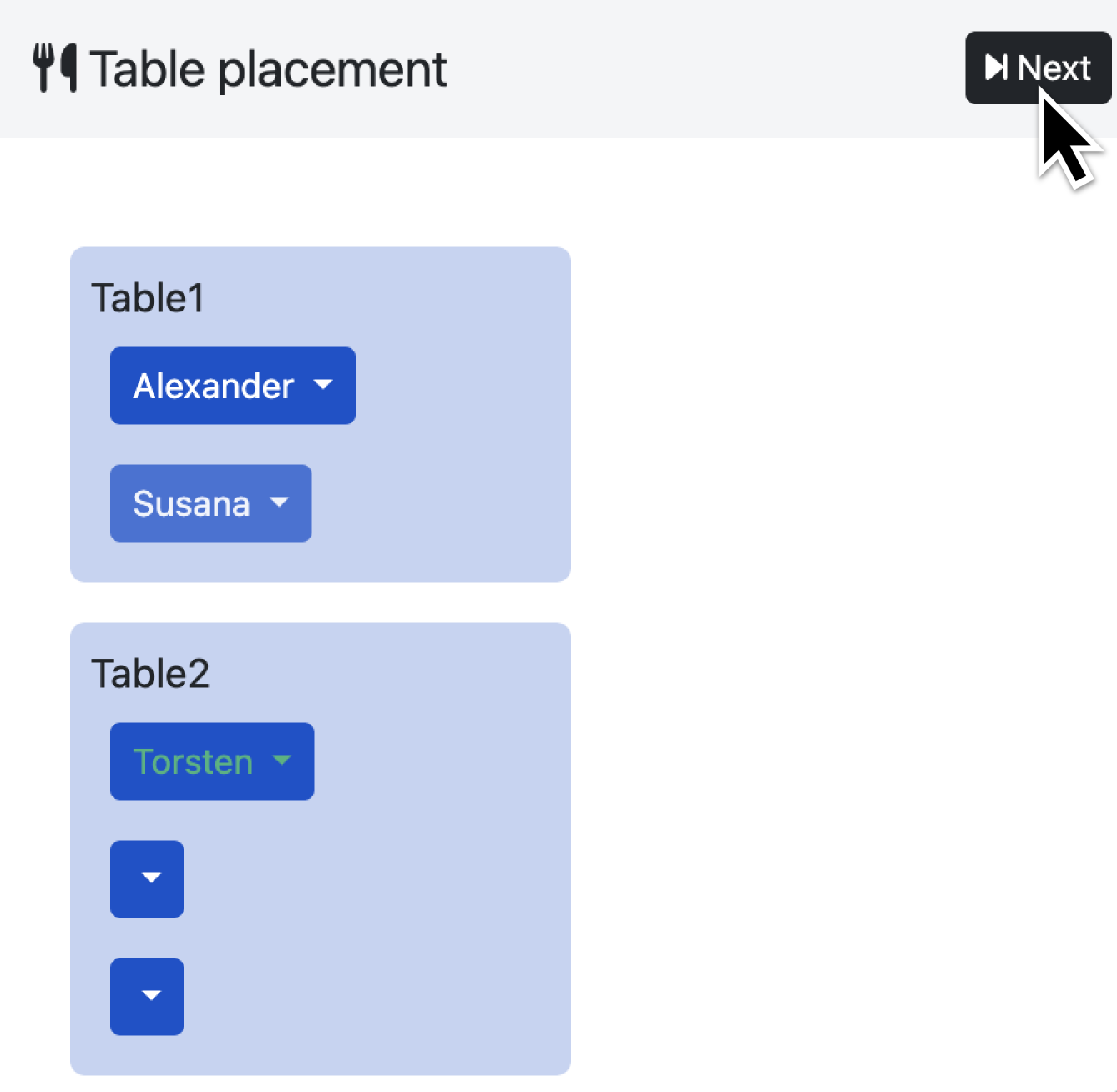}}
    \frame{\includegraphics[scale=0.20]{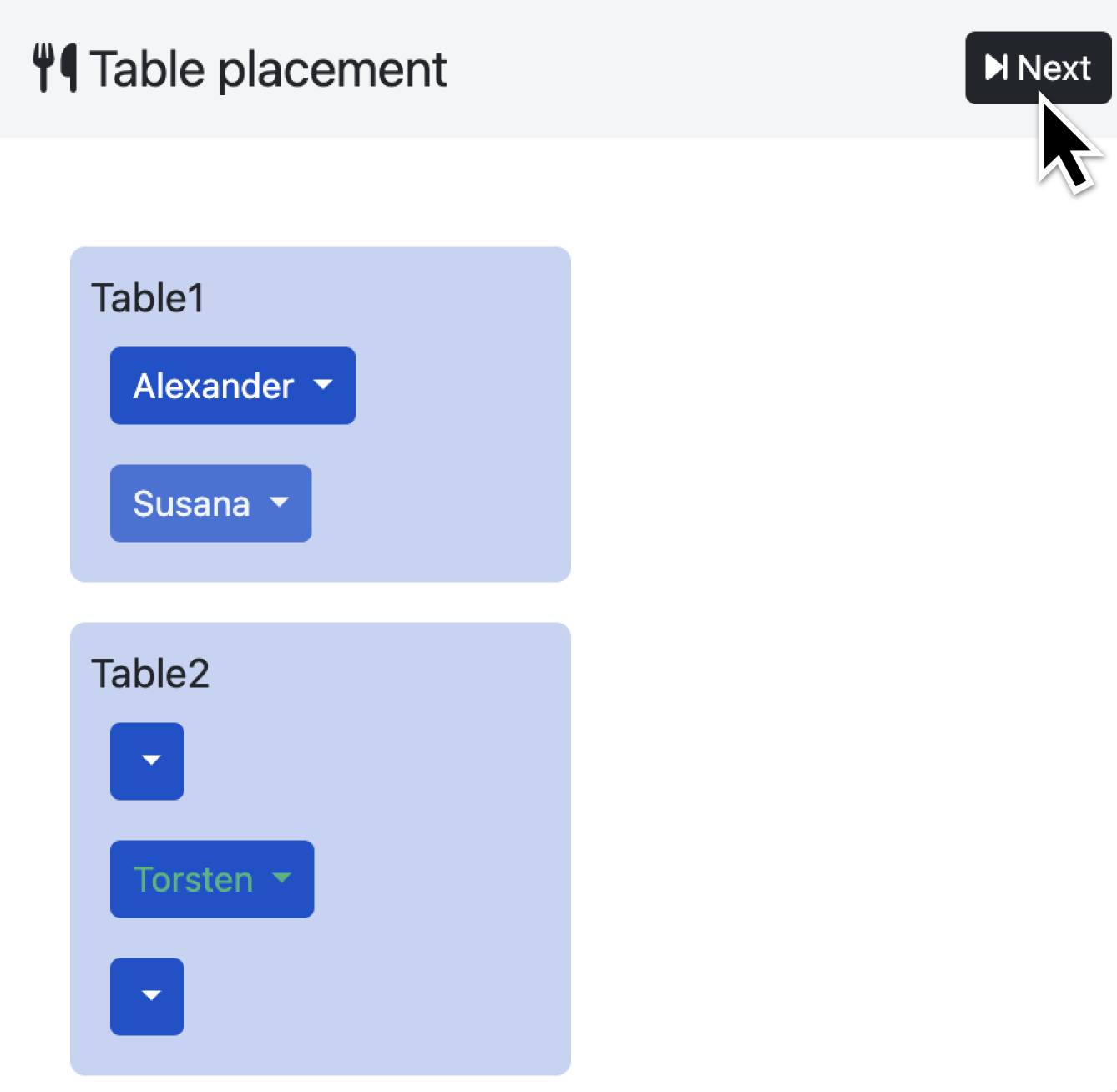}}
    \frame{\includegraphics[scale=0.20]{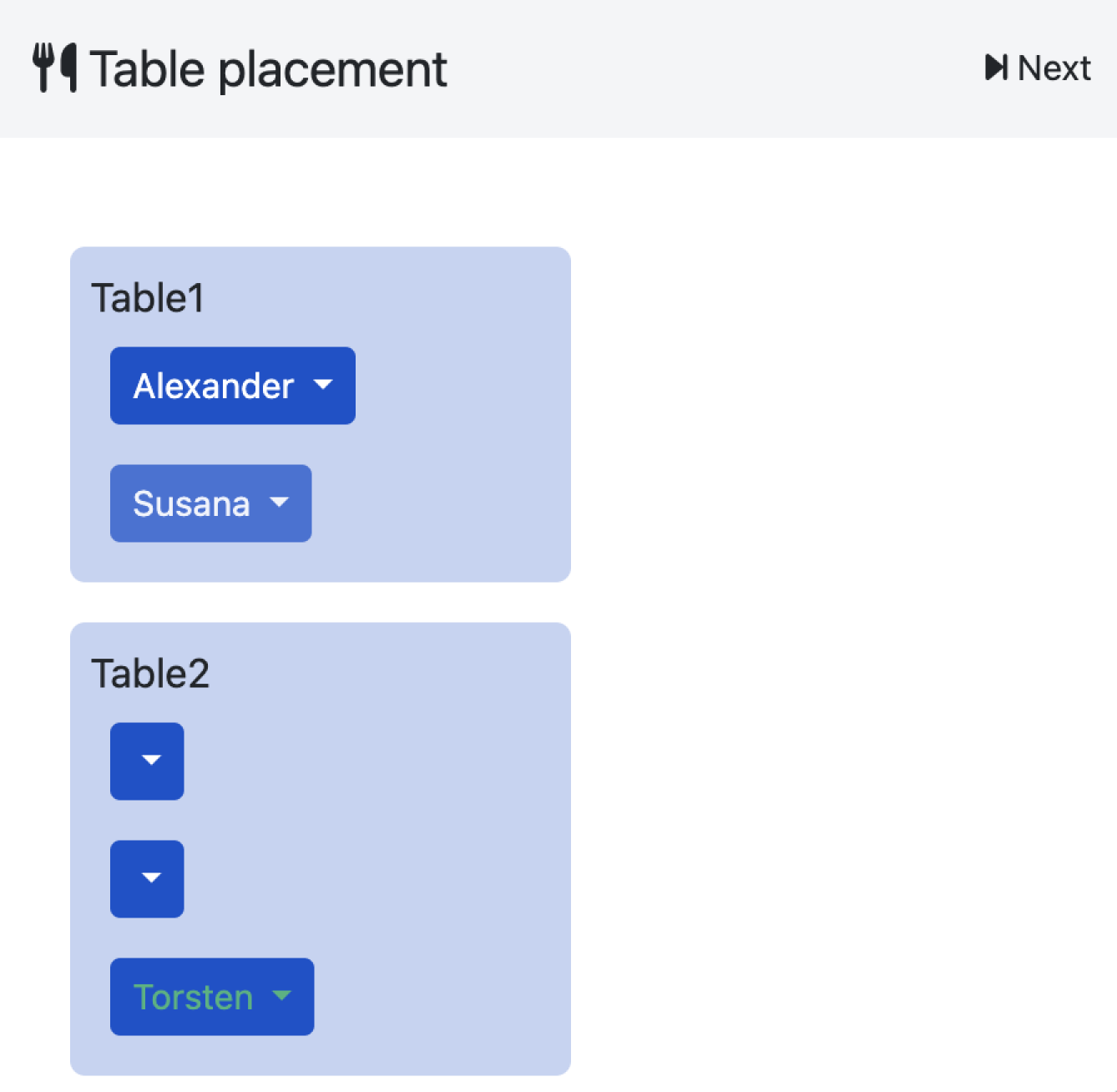}}
    \caption{User interaction via mouse actions using UI files \texttt{ui-tables.lp} and \texttt{ui-menu.lp}.}
    \label{fig:menu}
\end{figure}

\begin{lstlisting}[float=ht,caption={An encoding for menu handling (\texttt{ui-menu.lp})},label={prg:uimenu},language=clingos]
elem(menu_bar, menu_bar, w).%%#(\label{ui:menu:bar:one}#)
attr(menu_bar, title, "Table placement").%%#(\label{ui:menu:bar:two}#)
attr(menu_bar, icon, "fa-utensils").%%#(\label{ui:menu:bar:tri}#)

elem(menu_bar_next, button, menu_bar).%%#(\label{ui:menu:next:one}#)
attr(menu_bar_next, label, "Next").%%#(\label{ui:menu:next:two}#)
attr(menu_bar_next, icon, "fa-forward-step").%%#(\label{ui:menu:next:tri}#)
when(menu_bar_next, click, call, next_solution).%%#(\label{ui:menu:next:for}#)

attr(seat_dd(S), selected, P):- assign(P,S), _clinguin_browsing.%%#(\label{ui:menu:browsing:one}#)
attr(seat_dd(S), class, "text-success"):- _clinguin_browsing,%%#(\label{ui:menu:browsing:two}#)
                        assign(P,S), not _all(assign(P,S)).%%#(\label{ui:menu:browsing:tri}#)
attr(seat_dd(S), class, "opacity-75"):- _all(assign(P,S)),%%#(\label{ui:menu:browsing:for}#)
                        not _clinguin_assume(assign(P,S), true).%%#(\label{ui:menu:browsing:fiv}#)
\end{lstlisting}
For further illustration,
we now extend the functionality of our example by solution browsing.
That is, we add Listing~\ref{prg:uimenu} to Listing~\ref{prg:uitables} and
continue the user interaction from Figure~\ref{fig:table} in Figure~\ref{fig:menu}.
To begin with,
Lines~\ref{ui:menu:bar:one} to~\ref{ui:menu:bar:tri} create a \lstinline|menu_bar|
along with a \lstinline|title| and an \lstinline|icon|.\footnote{For handling icons, we use the icon library \emph{Font~Awesome} \url{https://fontawesome.com}.}
Lines~\ref{ui:menu:next:one} to~\ref{ui:menu:next:for} add a \lstinline|button| to the \lstinline|menu_bar| to iterate through solutions.
When the button is clicked, the server operation \lstinline|next_solution| is called.
This operation computes a new stable model for exploration and adds it to the \domainstate.

While the previous part of the UI encoding fixes static aspects of the user interface,
we now turn to the dynamic part.
Our design of solution browsing revolves around displaying the choices from the current stable model stored in the \domainstate.
This approach allows us to differentiate between three types of selections:
user selections,
derived necessary selections, on which all stable models agree, and
choices from the stable model under exploration.
To control this behavior, we rely upon the atom \lstinline|_clinguin_browsing|,
whose presence in the \domainstate\ indicates whether \clinguin\ is in browsing mode.
Recall that we show necessary selections in the UI by setting the \lstinline|selected| attribute of the dropdown menu
(in Line~\ref{ui:tables:dropdown:one:d} of Listing~\ref{prg:uitables}).
The same attribute is used in Line~\ref{ui:menu:browsing:one} to visualize choices of the stable model at hand,
when in browsing mode.
To visually distinguish the different choices,
we
reduce the opacity of derived necessary choices and
display non-necessary choices from the stable model at hand in green text;
user selections remain unchanged.
This is done in Lines~\ref{ui:menu:browsing:two} to~\ref{ui:menu:browsing:fiv}.
The first rule displays a seat assignment in green text, viz.\ \lstinline|"text-success"|,
if it belongs to the explored but not all stable models.
The second rule reduces the opacity of seat selections, which
belong to all stable models without being enforced by a user selection.
Since we represent user choices as assumptions,
we can easily determine if a seat selection originated from the user by verifying whether it was explicitly assumed or not.
The fact that \clinguin\ records all current assumptions in the \domainstate\
via predicate \lstinline|_clinguin_assume/2| in their reified form
explains the condition in Line~\ref{ui:menu:browsing:fiv}.

Figure~\ref{fig:menu} shows the effects of user interactions after performing the actions from Figure~\ref{fig:table}.
Unlike above,
the opacity of the dropdown for the second seat of Table~1 is now reduced,
indicating that this information is derived and not selected.
The user then clicks the Next button three times to iterate through all possible solutions.
These solutions show the three alternative seat assignments for Torsten in green.

\begin{figure}[ht]
    \centering
    \frame{\includegraphics[scale=0.20]{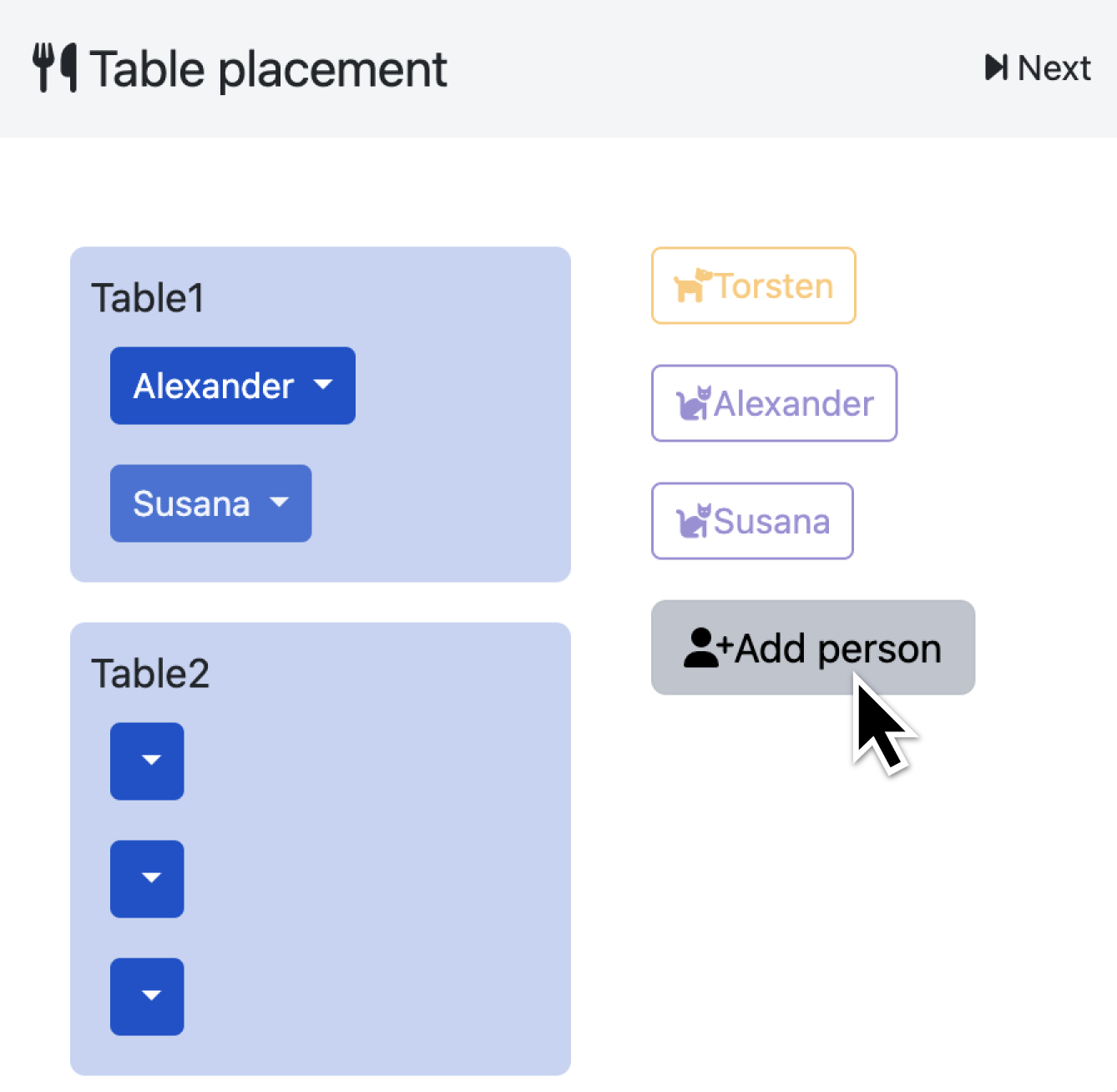}}
    \frame{\includegraphics[scale=0.20]{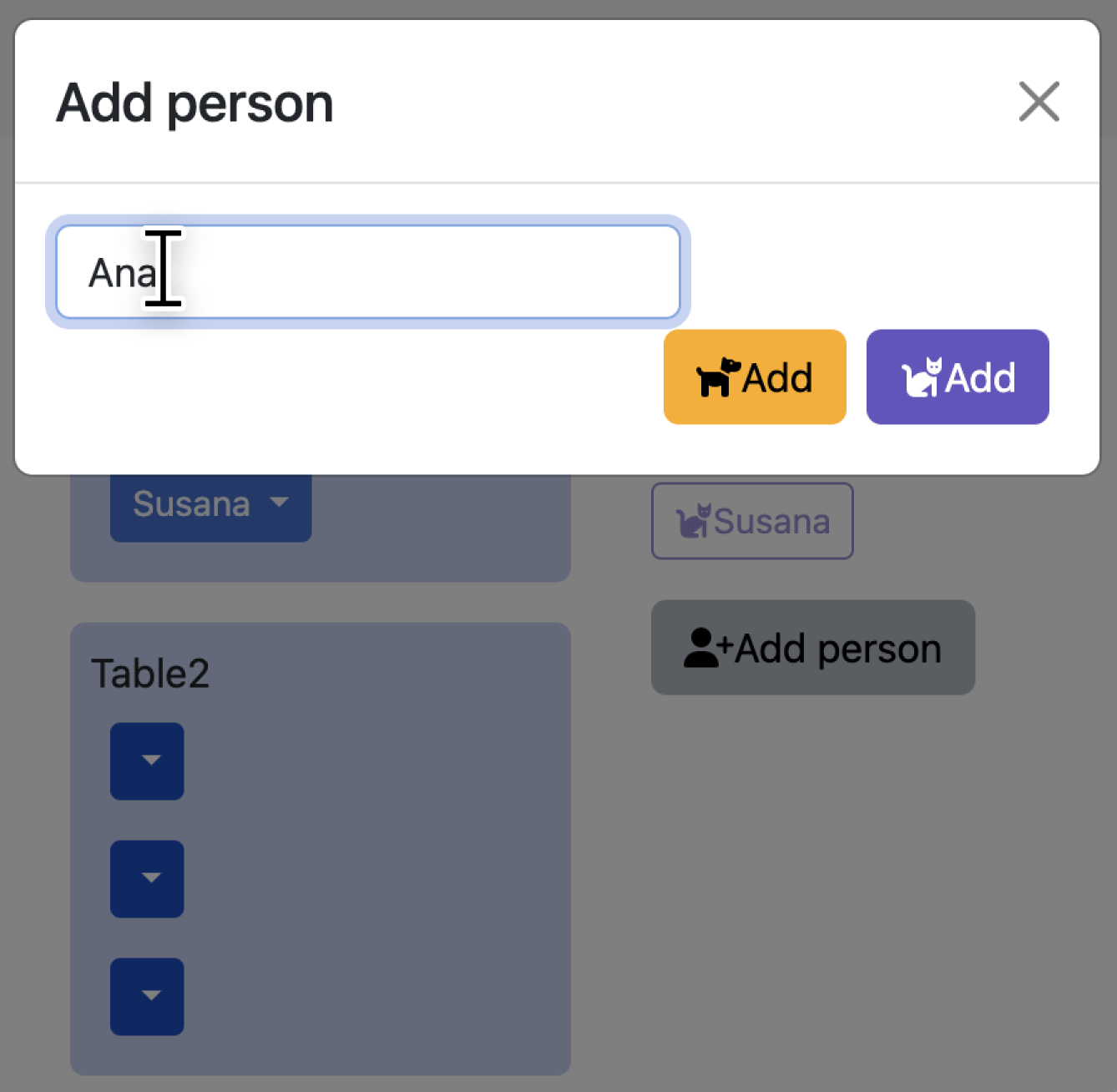}}
    \frame{\includegraphics[scale=0.20]{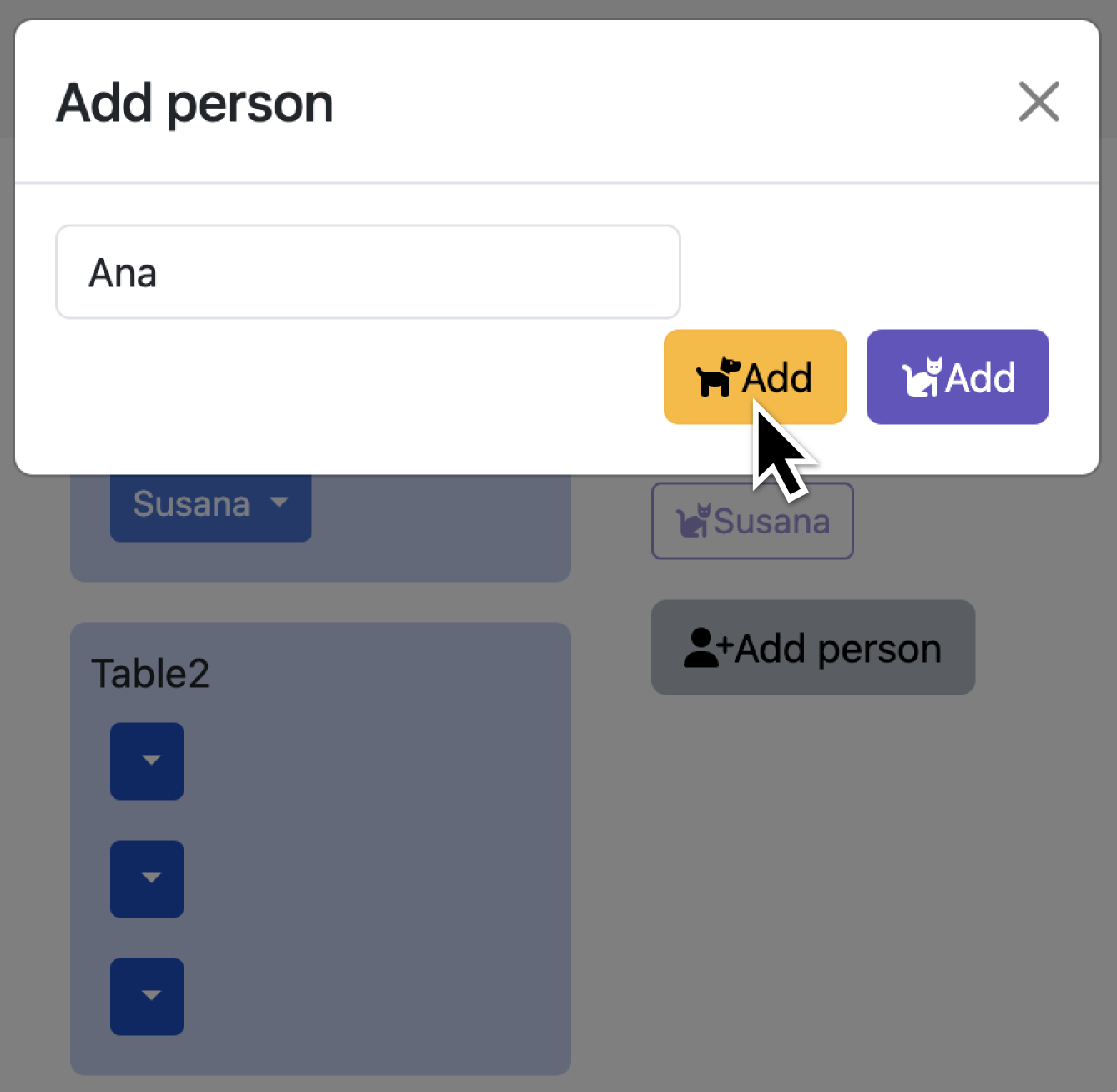}}
    \frame{\includegraphics[scale=0.20]{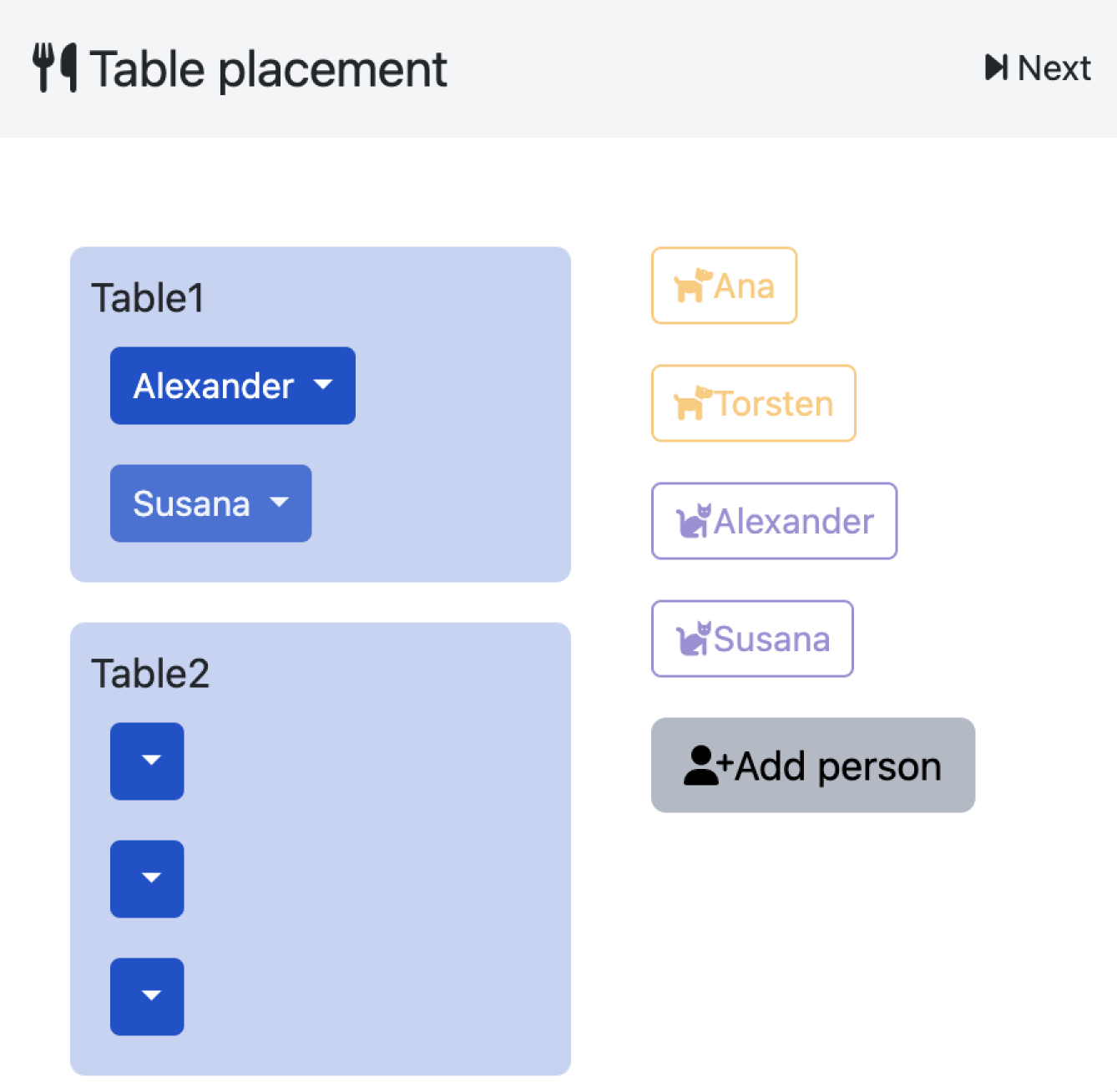}}
    \caption{User interaction via mouse actions using \texttt{ui-tables.lp}, \texttt{ui-menu.lp}, \texttt{ui-people.lp}.} \label{fig:people}
\end{figure}

Let us take our example a step further and explore how \clinguin\ incorporates user input using \emph{modals},
that is, UI elements that appear on top of the UI and temporarily deactivate the underlying content.
\begin{lstlisting}[caption={An encoding for instance generation (\texttt{ui-people.lp})},label={prg:uipeople},language=clingos]
elem(people, container, w).%%#(\label{ui:people:container}#)

elem(person(P), button, people):- person(P,_).%%#(\label{ui:people:person:one}#)
attr(person(P), label, P):- person(P,_).
attr(person(P), class, ("disabled";"m-2";"btn-sm")):- person(P,_).
attr(person(P), class, ("btn-outline-secondary")):- person(P,cat).
attr(person(P), class, ("btn-outline-warning")):- person(P,dog).
attr(person(P), icon, @concat("fa-",O)):- person(P,O).%%#(\label{ui:people:person:six}#)

elem(add_person, button, people).%%#(\label{ui:people:add:person:one}#)
attr(add_person, label, "Add person").
attr(add_person, icon, "fa-user-plus").
attr(add_person, class, ("btn-info";"m-2")).%%#(\label{ui:people:add:person:for}#)
when(add_person, click, update, (add_modal, visibility, shown)).%%#(\label{ui:people:add:person:fiv}#)

elem(add_modal, modal, w).%%#(\label{ui:people:modal:one}#)
attr(add_modal, title, "Add person").
elem(modal_content, container, add_modal).%%#(\label{ui:people:modal:container:one}#)
attr(modal_content, class, ("d-flex";"flex-column")).%%#(\label{ui:people:modal:container:two}#)

elem(name_tf, textfield, modal_content).%%#(\label{ui:people:modal:text:fields:one}#)
attr(name_tf, placeholder, "Enter the name").
attr(name_tf, order, 1).
attr(name_tf, width, 250).
when(name_tf, input, context, (name, _value)).%%#(\label{ui:people:modal:text:fields:fiv}#)

elem(btns_container, container, modal_content).%%#(\label{ui:people:buttons:container:one}#)
attr(btns_container, class, ("d-flex";"flex-row";"justify-content-end")).
attr(btns_container, order, 2).%%#(\label{ui:people:buttons:container:tri}#)

pet(cat;dog).%%#(\label{ui:people:buttons:one}#)
elem(add_btn(O), button, btns_container):- pet(O).
attr(add_btn(O), label, "Add"):- pet(O).
attr(add_btn(cat), class, ("m-1";"btn-secondary";"ml-auto")):- pet(O).
attr(add_btn(dog), class, ("m-1";"btn-warning";"ml-auto")):- pet(O).
attr(add_btn(O), icon, @concat("fa-",O)):- pet(O).
when(add_btn(O), click, context, (pet, O)):- pet(O).%%#(\label{ui:people:buttons:svn}#)
when(add_btn(O), click, call, %%#(\label{ui:people:buttons:eit}#)
     add_atom(person(_context_value(name,str), _context_value(pet)))):- pet(O).%%#(\label{ui:people:buttons:eitt}#)
\end{lstlisting}
More precisely,
we extend our UI encoding with Listing~\ref{prg:uipeople} to enable users to interactively
add people to
the problem instance,
as showcased in Figure~\ref{fig:people}.

To begin with, we create in Line~\ref{ui:people:container} a container to group all people in the instance.
In Lines~\ref{ui:people:person:one} to~\ref{ui:people:person:six} each person is represented as a disabled button,
whose color depends on the preferred pet.
We use buttons rather than labels to leverage on the icon functionality for representing pets.

Lines~\ref{ui:people:add:person:one} to~\ref{ui:people:add:person:for} define the button for adding new persons to the instance.
Once this button is clicked,
Line~\ref{ui:people:add:person:fiv} prescribes an \lstinline|update| operation
that pops up the modal element by
setting its \lstinline{visibility} attribute to \lstinline{shown}
(see the second snapshot in Figure~\ref{prg:uipeople}).
As shown in Figure~\ref{fig:clinguin},
this action is performed locally in the client without calling the server.

The actual modal element and its main container
are defined in Lines~\ref{ui:people:modal:one} to~\ref{ui:people:modal:container:two}.
The first element in the modal is a textfield for  user input;
it is defined in Lines~\ref{ui:people:modal:text:fields:one} to~\ref{ui:people:modal:text:fields:fiv}.
Whenever a user types an \lstinline|input| in the \lstinline|textfield|,
Line~\ref{ui:people:modal:text:fields:fiv} triggers the action \texttt{context},
which saves it in the dictionary of the same name in the client
(see Figure~\ref{fig:clinguin}).
The two parameters indicate that the key \lstinline{name} is assigned
the value of the input held by placeholder \lstinline|_value|.
Essentially, the user's input is assigned to a specific key within the dictionary.
The actual \emph{context} dictionary is stored locally on the client.
Any further user input in the textfield updates the value associated with the same key in the \emph{context} dictionary,
effectively replacing the previous input.
The entire \emph{context} dictionary is sent to the server along with every \lstinline|call| action triggered by the user.
Once the server sends a response, the \emph{context} dictionary is cleared on the client-side,
potentially to prepare for new interactions.

Lines~\ref{ui:people:buttons:container:one} to~\ref{ui:people:buttons:container:tri}
create a container for two additional buttons.
They are defined in Lines~\ref{ui:people:buttons:one} to~\ref{ui:people:buttons:eit},
one for adding cat persons and another for adding dog persons.
Most interesting is the reactivity of the buttons specified in
Lines~\ref{ui:people:buttons:svn} and~\ref{ui:people:buttons:eit}.
First of all, we note that a single event can trigger several actions.
Among them, local actions
are executed before \lstinline{call} actions.\footnote{
When multiple \lstinline{call} actions are triggered by the same event,
an order on the corresponding operations can be imposed
by including them in a tuple as the last argument of a single \lstinline|when| atom,
reflecting the desired order of execution.}
The action in Line~\ref{ui:people:buttons:svn} is
a local one and adds the selected \lstinline|pet| to the \emph{context} dictionary,
as described above for the \lstinline|name| entry.
Unlike this,
Lines~\ref{ui:people:buttons:eit} and \ref{ui:people:buttons:eitt} initiate a server \lstinline|call| adding atoms representing new persons,
using predicate \lstinline|person/2|.
The arguments for the atom are drawn from the \emph{context} dictionary.
This is accomplished via the lookup function \lstinline|_context_value(K)| which
yields the dictionary value for the key \lstinline|K|.
This replacement is done in the client and thus makes sure that the values are present before calling the server.
To provide further validation tools,
this lookup function allows for two optional arguments
indicating the expected type \lstinline{T} and a default value \lstinline|D|,
\lstinline|_context_value(K,T,D)|.
Possible values for types are \lstinline|str|, \lstinline|int| and \lstinline|const|
for strings, integers and terms, respectively.
By including a default value the presence of a value becomes optional.

Figure~\ref{prg:uipeople} shows how a new person can be added by a few clicks.
A click on the button `Add person' changes the visibility of the modal so that it appears in the window.
The user then types the person's name, viz.\ Ana, and clicks on the left button to add Ana as a dog person.
The last screenshot shows Ana as part of the people.
This is internally reflected by the addition of atom \lstinline|person("Ana",dog)| to the \domctl\ object.

While the above focuses on instance generation,
it's worth mentioning that \clinguin\ also allows users to download interactively entered problem instances.

 \section{Extensibility}
\label{sec:extensibility}

\Clinguin\ offers an open modular design that is easily extensible.
This allows us to implement extensions without changing the overall workflow.
On the server side, this is done by modifying the \textit{backend} component,
which provides a layer between the bare ASP solver and the user.
To achieve this extensibility,
we abstract the responsibilities of backends
into different key sections of \clinguin's workflow that can
be customized:
the set of callable operations,
the way the control object is handled (solving, grounding, model handling, etc.),
the atoms included in the \domainstate,
the way the UI is updated,
and the options passed when starting the server.
Currently, \clinguin\ offers a default backend using
\clingo, which implements all main functionalities for single- and multi-shot solving~\cite{gekakasc17a} via \clingo's API.
Additionally, it includes specialized backends for
\clingodl, a \clingo\ extension with difference constraints~\cite{jakaosscscwa17a},
\clingraph, a \clingo\ extension with visualization capabilities~\cite{hasascst22a},
and last but not least an explanation backend,
which we detail below.
On the client side, the \emph{frontend} component can be exchanged to accommodate different GUIs.
Currently, \clinguin\ offers the web-based frontend \emph{Angular}, used in the paper at hand, and
\emph{tkinter}, the standard \python\ interface to the \emph{Tcl/Tk GUI} toolkit.
Alternative GUI frameworks are easily incorporated,
mainly because the client-server communication of \clinguin\ is standardized by a JSON representation.

\subsection{Case study: A backend adding explanations}

This section focuses on extending \clinguin's backend to provide users with improved feedback
when their choices lead to an unsatisfiable scenario.
The first part
involves pinpointing specific conflicts within the user's selections that impede finding a solution.
The second one builds upon this by focusing on presenting error messages tailored to the specific reasons for unsatisfiability.

\begin{figure}[ht]
    \centering
    \frame{\includegraphics[scale=0.20]{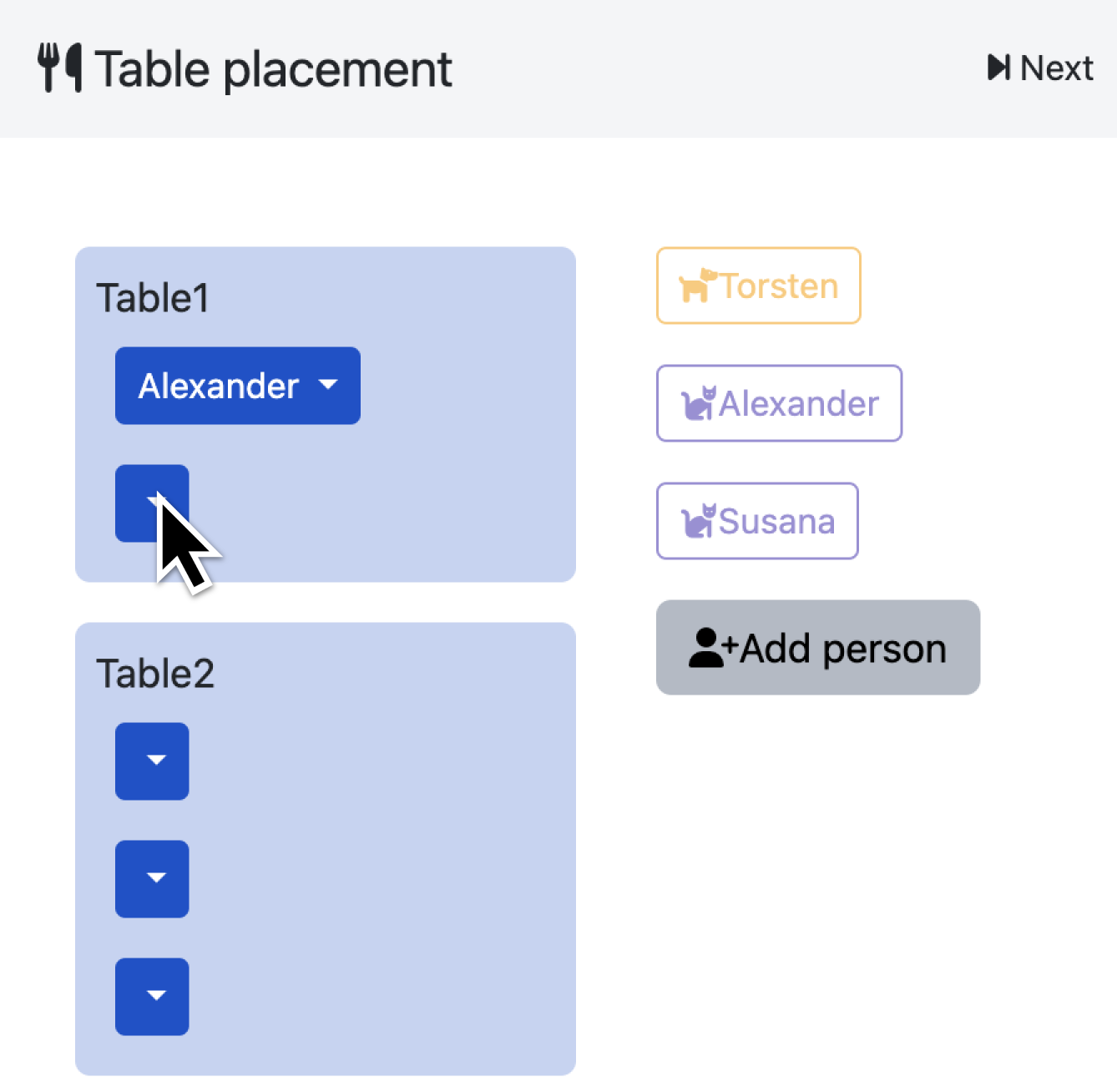}}
    \frame{\includegraphics[scale=0.20]{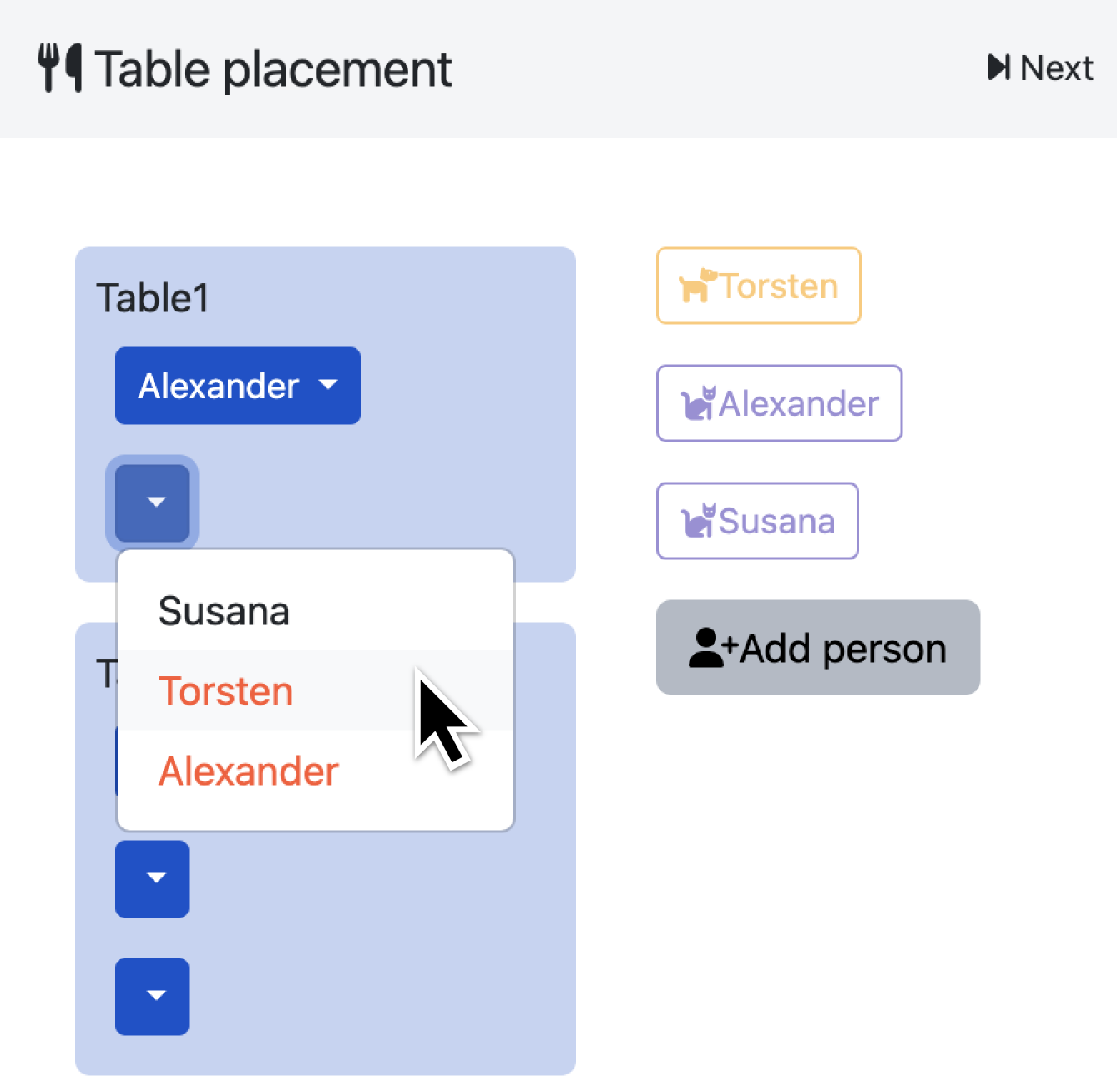}}
    \includegraphics[scale=0.20 ]{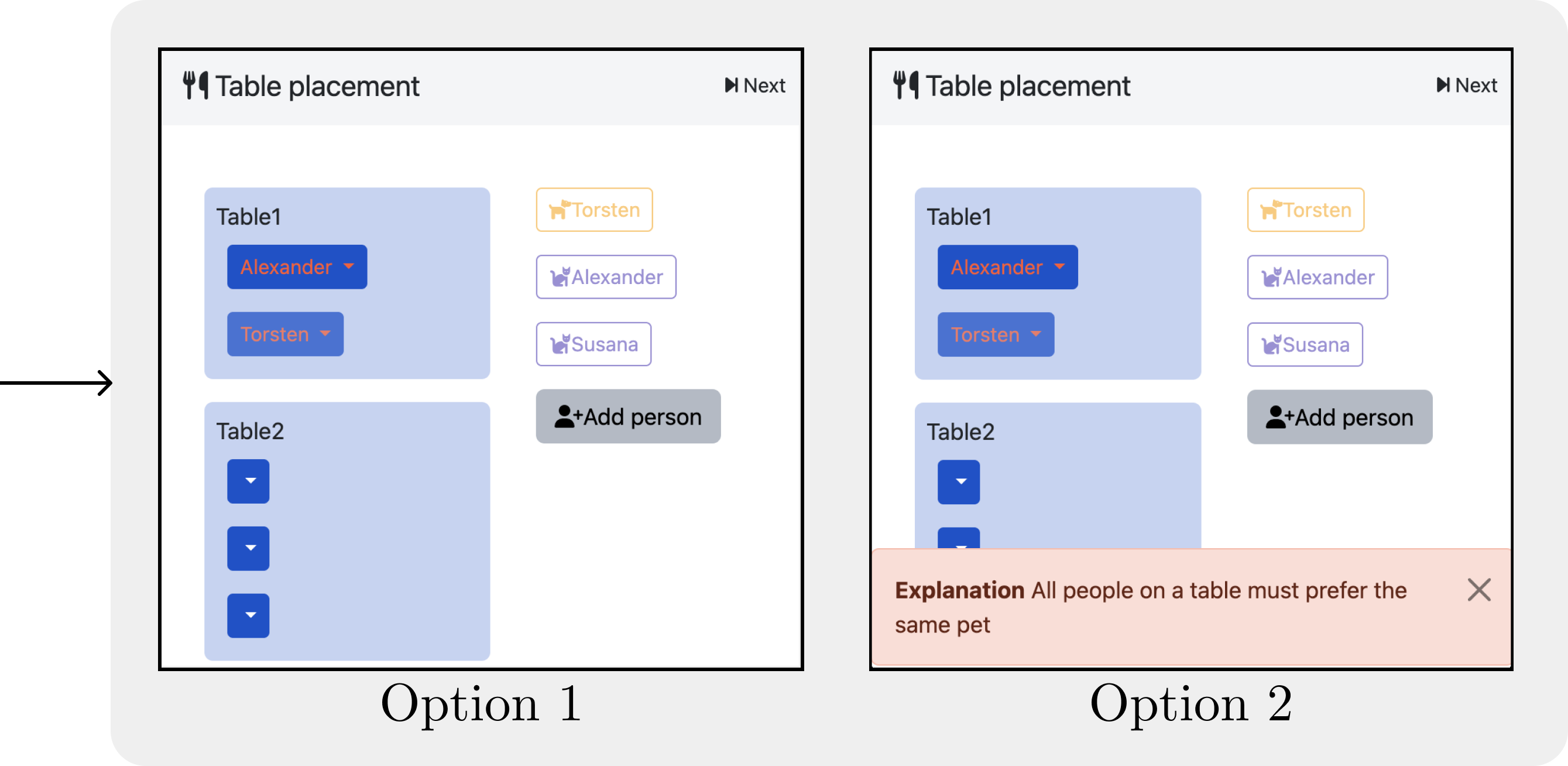}
    \caption{Example user interaction with the explanation extension.}
    \label{fig:explain}
\end{figure}

\begin{lstlisting}[caption={UI to handle basic explanations (\texttt{ui-explain.lp})},label={prg:uiexplain},language=clingos]
elem(seat_ddi(S,P), dropdown_menu_item, seat_dd(S)):-%%#(\label{ui:explain:one}#)
                          not _any(assign(P,S)), person(P,_), seat(S).
attr(seat_ddi(S,P), label, P):-%%#(\label{ui:explain:two}#)
                          not _any(assign(P,S)), person(P,_), seat(S).
when(seat_ddi(S,P), click, call, add_assumption(assign(P,S), true)):-%%#(\label{ui:explain:tri}#)
                          not _any(assign(P,S)), person(P,_), seat(S).%%#(\label{ui:explain:tri:b}#)
attr(seat_ddi(S,P), class, "text-danger"):-%%#(\label{ui:explain:for}#)
                          not _any(assign(P,S)), person(P,_), seat(S).
attr(seat_dd(S), class, ("text-danger")):- _clinguin_mus(assign(P,S)).%%#(\label{ui:explain:fiv}#)
\end{lstlisting}
We start by revealing all options, including those not leading to any solution.
This is done in Listing~\ref{prg:uiexplain} by exclusively using features explained above.
The encoding adds all infeasible options in Lines~\ref{ui:explain:one} to~\ref{ui:explain:tri:b}
and informs the user that they belong to no solution in Line~\ref{ui:explain:for}
(ignoring Line~\ref{ui:explain:fiv} for now).
The addition of \lstinline|dropdown-menu-item|[s] relies on the input from the standard backend.
Infeasible choices are indicated by red text.
Note the use of negative literals with predicate \lstinline{_any} to identify selections belonging to no stable model.
The effect of this approach is illustrated in the second snapshot of Figure~\ref{fig:explain}.

Our first approach hinges on identifying user choices that prevent a solution.
Here is
where the \emph{explanation backend} comes into play.
Remember that we model user selections as assumptions.
Modern ASP solvers like \clasp~\cite{ankamasc12a} or \wasp~\cite{aldofiprri23a}
can pinpoint unsatisfiable sets of assumptions when solving an ASP program.
We exploit this capability in \clinguin's explanation backend
by including instances of the special predicate \lstinline{_clinguin_mus/1} in the \domainstate.
The arguments of all such instances represent a minimal\footnote{The set is minimal in the sense that removing any of its assumptions leads to a satisfiable result.}
set of assumptions that, when combined, lead to an unsatisfiable scenario.

\Clinguin's \lstinline{_clinguin_mus/1} predicate helps us pinpoint infeasible user choices.
We leverage this predicate in Line~\ref{ui:explain:fiv}
to highlight in red the \lstinline|dropdown_menu|[s] of the assignments causing the issue
(extending Lines~\ref{ui:tables:dropdown:one:a} to~\ref{ui:tables:dropdown:one:d} in Listing~\ref{prg:uitables}).
Additionally, to only show the selection in menus associated with user choices,
we modify Line~\ref{ui:tables:dropdown:one:d} in Listing~\ref{prg:uitables},
and replace predicate \lstinline{_all} with \lstinline{_clinguin_assume} in the body of the rule:
\begin{lstlisting}[firstnumber=21,language=clingos]
attr(seat_dd(S), selected, P):- _clinguin_assume(assign(P,S),true).
\end{lstlisting}
This guarantees that the text in the dropdown menus reflects the user's selection,
even if it leads to an unsatisfiable result.
When an assumption causes an unsatisfiable scenario,
the facts in the \domainstate\ coming from the \domctl\ (including the stable model, \lstinline{_any}, and \lstinline{_all})
are retrieved from the last successful computation.
Meanwhile, those indicating the state of unsatisfiability and selections (given by \lstinline{_clinguin_mus/1} and \lstinline{_clinguin_assume/2})
are effectively updated.
Therefore, in this extension handling unsatisfiable scenarios,
using predicate \lstinline{_clinguin_assume} instead of \lstinline{_all}
ensures the selection is accurately displayed.

An interaction using this UI feature is shown in Figure~\ref{fig:explain}, ending in Option~1.
As before, we start by selecting Alexander for the first seat at Table~1.
However, unlike Figure~\ref{fig:table}--\ref{fig:people},
the second seat remains unassigned due to the code modification in Listing~\ref{prg:uitables}.
Clicking on this reveals both Torsten and Alexander highlighted in red.
This signifies that neither option leads to a valid solution.
Now, imagine a curious user selects Torsten despite the red color,
aiming to understand why this choice is infeasible.
As a result, the dropdown menus of the specific assumptions causing the unsatisfiability are highlighted in red.
In this simple example, it might only indicate a conflict with Alexander's previous assignment.
However, in more complex scenarios, only a subset of assumptions would typically be highlighted,
pinpointing the exact source of the infeasibility.

The interaction leading to Option~1 in Figure~\ref{fig:explain} is obtained by the following command:
\begin{lstlisting}[basicstyle=\small\ttfamily,numbers=none]
 clinguin client-server --domain-files ins.lp enc.lp \
 --ui-files ui-tables.lp ui-menu.lp ui-people.lp ui-explain.lp \
 --backend ExplanationBackend
\end{lstlisting}
This command
adds \lstinline|ui-explain.lp| and the changed \lstinline|ui-tables.lp| file to the UI files and
moreover an additional parameter to indicate the use backend \lstinline|ExplanationBackend|.

Our second part of the extension aims at providing user-friendly explanations in natural language when their choices lead to dead ends.
For instance, in our running example,
Alexander and Torsten cannot be placed at the same table because they have conflicting pet preferences.
Ideally,
the UI should display an error message that clearly explains this specific reason whenever both are assigned to the same table.
The key to generating these explanations lies in their connection to the domain files.
Since explanations are specific to the problem being modeled,
we leverage the concept of integrity constraints defined in those files.
By tying unsatisfiability to the violation of these constraints,
we can precraft corresponding explanations that pinpoint the exact reason behind the dead end.

To this end, we use the binary predicate \lstinline{cons/2} in Listing~\ref{prg:encoding} to identify and trigger constraints.
In essence, each integrity constraint is paired with a \lstinline{cons/2} instance that acts as a translator,
converting the technical constraint violation into an understandable explanation for the user.
In order to have these atoms as part of our minimal set of assumptions,
these backend extension performs a transformation of the domain files.
This transformation will (internally) replace facts matching the predicate signature \lstinline{cons/2} by a choice,
providing the option to activate or deactivate the corresponding constraint when searching for unsatisfiable assumptions.
Then, we treat each instance of \lstinline{cons/2} as a true assumption,
which keeps all original integrity constraints intact.

Now that we understand how \lstinline{cons/2} instances link constraints to explanations,
let us explore how these explanations are triggered within the system.
Whenever a specific \lstinline{cons/2} instance ends up in an unsatisfiable set of atoms,
it indicates that the constraint must be active to trigger the inconsistency,
which can be interpreted as a violation of the corresponding constraint.
This violation triggers the explanation associated with that \lstinline{cons/2} instance in the UI
by opening a \lstinline{message} box displaying the explanation.
This is defined in Listing~\ref{prg:uiexplainmsg} for \lstinline{cons/2} instances belonging to the current set of
assumptions held in the server's \domainstate.
\begin{lstlisting}[caption={UI to handle complex explanations (\texttt{ui-explain-msg.lp})},label={prg:uiexplainmsg},language=clingos]
elem(message_unsat(N), message, w):-_clinguin_mus(cons(N,M)).
attr(message_unsat(N), title, "Explanation"):-_clinguin_mus(cons(N,M)).
attr(message_unsat(N), message, M):-_clinguin_mus(cons(N,M)).
attr(message_unsat(N), type, error):-_clinguin_mus(cons(N,M)).
\end{lstlisting}

Finally,
we can engage the interaction leading to Option~2 in Figure~\ref{fig:explain} as follows:
\begin{lstlisting}[basicstyle=\footnotesize\ttfamily,numbers=none]
 clinguin client-server --domain-files ins.lp enc.lp \
 --ui-files ui-tables.lp ui-menu.lp ui-people.lp \
            ui-explain.lp ui-explain-msg.lp \
 --backend ExplanationBackend --assumption-signature cons,2
\end{lstlisting}
We add \texttt{ui-explain-msg.lp} to the previous set of UI files,
and indicate the spacial treatment of facts matching a predicate signature
by using the command line option \lstinline{--assumption-signature} in a generic way.

We now address the implementation details of this \clinguin\ backend.
The \texttt{ExplanationBackend} extends the available standard backend for \clingo, viz \texttt{ClingoBackend}.
In detail,
it alters the setup to keep track of the internal representation of the assumptions,
registers a new command line option \texttt{assumption-signature}, and
adds (minimally) unsatisfiable sets of assumptions to the \domainstate.

Listing~\ref{prg:explanationbackend} presents the complete \python\ implementation of the \texttt{ExplanationBackend} class.
\begin{lstlisting}[float=h, caption={Python implementation of \texttt{ExplanationBackend}},label={prg:explanationbackend},language=python, basicstyle=\footnotesize\ttfamily]
class ExplanationBackend(ClingoBackend):

    @classmethod
    def register_options(cls, parser): #( \label{prg:explanation:backend:ro:begin} #)
        ClingoBackend.register_options(parser)

        parser.add_argument(
            "--assumption-signature",
            help="Signatures that will be considered as true assumptions",
            nargs="+") #( \label{prg:explanation:backend:ro:end} #)
        
    def _init_command_line(self): #( \label{prg:explanation:backend:initc:begin} #)
        super()._init_command_line() #( \label{prg:explanation:backend:initc:super} #)
        self._assumption_sig = [] #( \label{prg:explanation:backend:initc:as} #)
        for a in self._args.assumption_signature or []:
            name, arity = a.split(",") #( \label{prg:explanation:backend:initc:lop} #)
            self._assumption_sig.append((name, int(arity)))
        self._assumption_transformer = AssumptionTransformer(self._assumption_sig) #( \label{prg:explanation:backend:initc:at} #)

    @property
    def _assumption_list(self): #( \label{prg:explanation:backend:as:begin} #)
        return self._assumptions.union(self._assumptions_from_signature) #( \label{prg:explanation:backend:as:end} #)

    def _load_file(self, f): #( \label{prg:explanation:backend:lf:begin} #)
        transformed_program = self._assumption_transformer.parse_files([f])
        self._ctl.add("base", [], transformed_program) #( \label{prg:explanation:backend:lf:end} #)

    def _ground(self, program="base", arguments=None): #( \label{prg:explanation:backend:g:begin} #)
        super()._ground(program, arguments)
        transformer_assumptions = self._assumption_transformer.get_assumption_symbols(
            self._ctl, self._ctl_arguments_list)
        self._assumptions_from_signature = [(a, True) for a in transformer_assumptions] #( \label{prg:explanation:backend:g:end} #)

    def _init_ds_constructors(self): #( \label{prg:explanation:backend:initd:begin} #)
        super()._init_ds_constructors()
        self._add_domain_state_constructor("_ds_mus") #( \label{prg:explanation:backend:initd:end} #)

    @cached_property
    def _ds_mus(self): #( \label{prg:explanation:backend:ds:begin} #)
        prg = "#defined _clinguin_mus/1. " #( \label{prg:explanation:backend:ds:defined} #)
        if self._unsat_core is not None: #( \label{prg:explanation:backend:ds:mus:mus:begin} #)
            mus_core = CoreComputer(self._ctl, self._assumption_list).shrink() #( \label{prg:explanation:backend:ro:core} #)
            prg += " ".join([f"_clinguin_mus({str(s)})." for s, _ in mus_core])#( \label{prg:explanation:backend:ds:mus:mus:end} #)
        return prg + "\n" #( \label{prg:explanation:backend:ds:end} #)
\end{lstlisting}
The \lstinline|classmethod| in
Lines~\ref{prg:explanation:backend:ro:begin} to~\ref{prg:explanation:backend:ro:end}
adds the \lstinline|--assumption-signature| argument to the command line.
Its value is then processed in the method \lstinline|_init_command_line|
to extract the predicate signatures,
as shown in Lines~\ref{prg:explanation:backend:initc:begin} to~\ref{prg:explanation:backend:initc:at}.
Of special interest is Line~\ref{prg:explanation:backend:initc:at}, which creates an object
of the class \lstinline|AssumptionTransformer|, defined by the Python library \lstinline|clingexplaid|
\footnote{
    \url{https://github.com/potassco/clingo-explaid}
}.
This library provides the necessary functionalities to transform the domain files
and minimize the set of unsatisfiable assumptions.
The transformer is then used in the \lstinline|_load_file| method
(Lines~\ref{prg:explanation:backend:lf:begin} to~\ref{prg:explanation:backend:lf:end})
to change how files are loaded into the \domainstate\ in the standard backend.
Specifically, it transforms the facts in these files into choices based on the given predicate signatures.
This object is also used after grounding (Lines~\ref{prg:explanation:backend:g:begin} to~\ref{prg:explanation:backend:g:end})
to retrieve the list of assumptions from the predicate signature and store them.
In Lines~\ref{prg:explanation:backend:as:begin} to~\ref{prg:explanation:backend:as:end}, these assumptions
are included in the property \lstinline|_assumption_list|, which is used as a parameter when solving.
Finally, the domain constructor in
Lines~\ref{prg:explanation:backend:ds:begin} to~\ref{prg:explanation:backend:ds:end}
returns a program that gets added to the \domainstate\
by the method \lstinline|_init_ds_constructors| in Lines~\ref{prg:explanation:backend:initd:begin} to~\ref{prg:explanation:backend:initd:end}.
Line~\ref{prg:explanation:backend:ds:defined} adds a \lstinline|#defined| statement for the predicate \lstinline|_clinguin_mus/1|
to avoid warnings.
The code in Lines~\ref{prg:explanation:backend:ds:mus:mus:begin} to~\ref{prg:explanation:backend:ds:mus:mus:end}
is executed only when the output of the \domctl\ is unsatisfiable.
In this case,
it retrieves a minimal set of unsatisfiable assumptions
using an object of the class \lstinline|CoreComputer| from the same library,
which is added to the \domainstate\ (Line~\ref{prg:explanation:backend:ds:mus:mus:end}).
The decorator \lstinline|@cached_property| ensures efficient computation by reusing results unless the cache is invalidated.

 \section{Related work}
\label{sec:related}

The creation of problem-specific interfaces has been a long standing challenge for declarative methods.
A \prolog-based method using XML was investigated in~\cite{schsei12a}.
\cite{schel96a} focus on automatic user interface generation with model-based UIs.
This is extended with contextual information and ASP in~\cite{zakzag11a}.
In the context of non-interactive ASP visualizations, recent advancements have led to \clingraph~\cite{hasascst22a}.
The need for interactivity in ASP was addressed in previous studies~\cite{geobsc15a},
and later incorporated into \clingo's API with multi-shot capabilities~\cite{karoscwa21a},
enabling continuous solving of logic programs that undergo frequent changes.
Tools have also emerged to facilitate ASP program development,
such as ASP Chef~\cite{alcrro23a} for task pipelining,
and various Integrated Development Environments~\cite{fereri11a,buoepuskto13a}.
These tools align with advancements in related areas like argumentation~\cite{dachselt2022nexas}.
In contrast to existing work,
\clinguin\ focuses on creating modern domain-specific interactive user interfaces in ASP.

\section{Conclusion}\label{sec:discussion}

We have presented \clinguin, an easy yet expressive tool for creating user interfaces within ASP.
This application relies on ASP features, like
union and intersection of stable models,
assumptions, and
minimally unsatisfiable sets.
A central idea is to reify these concepts and to keep them along with a stable model in focus as a set of atoms,
which is then used by a UI encoding to generate a user interface and react to user events.
Meanwhile,
\clinguin\ has become invaluable in our industrial applications
since it greatly extends the rapid prototyping nature of ASP.
\Clinguin\ (version \lstinline|2.0|) is freely available as open-source software at \url{https://github.com/potassco/clinguin};
its documentation is obtained at \url{https://clinguin.readthedocs.io}.
The distribution contains several substantial use-cases, illustrating the full power of \clinguin.

\bibliographystyle{eptcs}

\end{document}